\documentclass[11pt]{article}

\usepackage[final]{acl}

\usepackage{times}
\usepackage{latexsym}

\usepackage[T1]{fontenc}
\usepackage[utf8]{inputenc}

\usepackage{microtype}

\usepackage{inconsolata}

\usepackage{graphicx}
\usepackage{amsmath}
\usepackage{amssymb}
\usepackage{booktabs}
\usepackage{multirow}
\usepackage{pifont}
\usepackage{algorithm}
\usepackage{algpseudocode}
\usepackage{float}
\usepackage[normalem]{ulem}
\useunder{\uline}{\ul}{}
\usepackage[most]{tcolorbox}
\usepackage{xcolor}
\usepackage{makecell}
\usepackage{enumitem}

\usepackage{titlesec}
\titlespacing*{\section}{0pt}{10pt}{4pt}
\titlespacing*{\subsection}{0pt}{8pt}{3pt}
\titlespacing*{\subsubsection}{0pt}{6pt}{2pt}
\titlespacing*{\paragraph}{0pt}{4pt}{0.5em}
\titlespacing*{\subparagraph}{0pt}{3pt}{0.5em}
\usepackage{enumitem}
\setlist[itemize]{topsep=2pt, partopsep=0pt, itemsep=2pt, parsep=2pt, leftmargin=*}
\setlist[enumerate]{topsep=2pt, partopsep=0pt, itemsep=2pt, parsep=2pt, leftmargin=*}
\title{Plan-and-Verify Video Reward Reasoning with Spatio-Temporal Scene Graph Grounding}

\author{
  \textbf{Hyomin Kim\textsuperscript{1}\textsuperscript{*}},
  \textbf{Junghye Kim\textsuperscript{1}\textsuperscript{*}},
  \textbf{Joanie Hayoun Chung\textsuperscript{2}},
  \textbf{Yoonjin Oh\textsuperscript{1}},
  \textbf{Kyungjae Lee\textsuperscript{2}},
\\
  \textbf{Sungbin Lim\textsuperscript{2}\textsuperscript{\textdagger}},
  \textbf{Sungwoong Kim\textsuperscript{1}\textsuperscript{\textdagger}}
\\
\\
  \textsuperscript{1}Department of Artificial Intelligence, Korea University,
  \textsuperscript{2}Department of Statistics, Korea University
\\
  \small{
    \textsuperscript{*}Equal contribution\quad 
    \textsuperscript{\textdagger}Corresponding authors
  }
\\
  \small{
    \{khmiee, jh0912, sungbin, swkim\}@korea.ac.kr
  }
}

\begin{document}
\maketitle

\begin{abstract}
Reward models for text-to-video (T2V) generation guide post-training but often fail at fine-grained semantic alignment.
We trace this to two structural weaknesses in existing reasoning-based reward models: they do not systematically verify every condition described in the prompt, and the visual evidence supporting each judgment remains implicit in their free-form reasoning. 
We propose \textbf{SG-PVR}, a video reward model that addresses these limitations through plan-and-verify reasoning grounded in spatio-temporal scene graphs. 
The verification plan decomposes the prompt into atomic claims, making the set of requirements to be checked explicit.
The spatio-temporal scene graph, encoding entities, attributes, and temporally grounded relations, is extracted from the video and maintained as a persistent structured visual reference throughout reasoning.
Each claim is verified against both the video and the scene graph, anchoring judgments in explicit visual evidence.
SG-PVR achieves strong performance on semantic alignment, including fine-grained temporal semantics.
As a test-time reranker, it further enhances compositional alignment in T2V generation.
\end{abstract}

\section{Introduction}
\label{sec:introduction}
Text-to-video (T2V) generation has advanced with recent diffusion transformer-based models producing increasingly realistic, high-resolution videos~\cite{wan2025wan, kong2024hunyuanvideo, seedance2026seedance}. 
To close the remaining gap to human preference, post-training methods such as Direct Preference Optimization (DPO)~\cite{liu2025videodpo, liu2025improving} and Group Relative Policy Optimization (GRPO)~\cite{xue2025dancegrpo, liu2025flow} align T2V models with preference judgments. 
The reliability of any such procedure depends on the reward signal: violations that the reward model (RM) fails to detect are unlikely to be corrected.
Since T2V prompts often contain multiple entities, attributes, actions, and temporal relations, the RM must identify which requirements are satisfied or violated rather than assigning one global preference score.
Designing a video RM that captures fine-grained, compositional alignment between a prompt and a generated video is therefore a central bottleneck for further T2V quality gains.

Existing video RMs, however, fall short on this fine-grained semantic alignment. 
Non-reasoning RMs produce single scalar or per-dimension scores~\cite{liu2025improving,he2024videoscore,wang2025unified}; these scores conflate spatial, temporal, and attribute errors.
Reasoning RMs improve interpretability by generating chain-of-thought (CoT) rationales before scoring~\cite{he2025videoscore2,wang2025unifiedthink}, but suffer from two complementary weaknesses: 
% (i) lack of coverage, where free-form rationales only judge the claims the model elects to verbalize, leaving the remaining prompt requirements unverified; 
(i) \emph{unspecified verification targets}, where the set of prompt requirements to be evaluated is not explicitly specified, making it unclear whether all necessary prompt elements have been checked;
(ii) \emph{weak complex-temporal alignment}, where video-only free-form reasoning can struggle to evaluate alignment for prompts involving event order, state changes, or complex relations.

We propose \textbf{SG-PVR}, a video reward reasoning framework with two components addressing the limitations above.
To address (i), SG-PVR adopts a plan-and-verify structure: the model first decomposes the prompt into atomic verification claims, explicitly specifying what should be checked before performing per-claim verification.
%To address (i), we adopt a plan-and-verify structure, in which the model first decomposes the prompt into a verification plan of atomic claims, then verifies each claim and aggregates the outcomes---ensuring no prompt requirement is left unverified. 
To address (ii), SG-PVR represents the video as a spatio-temporal scene graph, providing structured visual evidence of entities, attributes, and temporal relations for complex-temporal alignment evaluation.
%To address (ii), we make the visual evidence used at each verification step explicit and structured, instantiating it as a spatio-temporal scene graph of entities, attributes, and temporally-grounded relations extracted from the video and kept in the reasoning context throughout the trace. 

During alignment evaluation, SG-PVR jointly uses the raw video and the scene graph: the video provides the original visual evidence, while the graph organizes temporal relations and event-level information that is difficult to track with video-only reasoning.
This improves alignment evaluation for prompts involving event order, state changes, and complex relations.
SG-PVR also evaluates video quality within the same reasoning trace, producing both semantic alignment and quality scores in a single generation.
%The graph is not treated as a definitive source. 
%Instead, each claim is verified against both the graph and the raw video, so that the video can provide evidence missing from the graph while also correcting relations that may be hallucinated.
%Beyond semantic alignment, the model simultaneously evaluates intrinsic video quality within the same reasoning trace, producing both semantic and quality scores in a single generation.

Our contributions are as follows. 
\begin{itemize}
\item \textbf{SG-grounded reasoning for video reward modeling.} 
We introduce a spatio-temporal scene graph as a structured intermediate representation that persists in the reasoning context of a video RM.
The graph is complementary to raw video evidence, with each verification consulting both.
\item \textbf{Plan-and-verify reasoning.} 
Rather than free-form CoT, we structure semantic alignment evaluation as decomposing the prompt into atomic claims, verifying each against the video and the scene graph, and aggregating per-claim outcomes through a rubric-guided summarization.
% This ensures that all prompt requirements are covered, each judgment is tied to explicit evidence, and aggregation reflects the structure of semantic failures rather than averaging them uniformly.
This makes prompt requirements explicit, ties each judgment to explicit evidence, and allows aggregation to reflect the structure of semantic failures rather than averaging them uniformly.
\item \textbf{Strong reasoning RM for T2V evaluation.}
SG-PVR demonstrates strong semantic alignment across pointwise reward benchmarks and achieves the highest overall accuracy on fine-grained temporal semantics.
As a test-time reranker, it improves compositional alignment in T2V generation.
\end{itemize}

\begin{figure*}[!t]
    \centering
    \includegraphics[width=\linewidth]{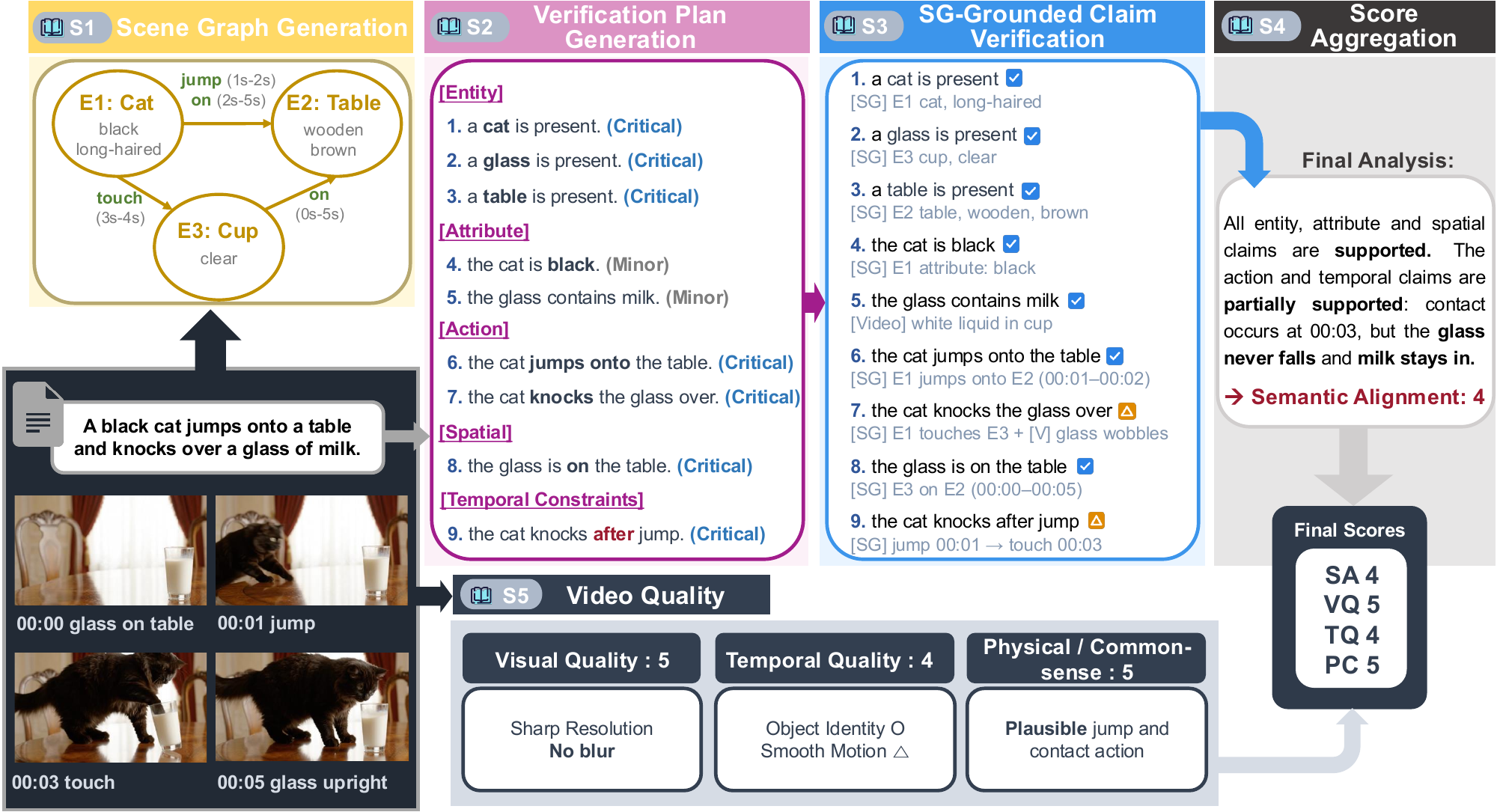}
    \caption{
    Overview of SG-PVR. 
    Given a prompt $P$ and video $V$, SG-PVR (i) extracts a spatio-temporal scene graph $\mathcal{G}$ from $V$, (ii) decomposes $P$ into a verification plan of atomic claims tagged Critical or Minor, (iii) verifies each claim against $\mathcal{G}$ and $V$ as Supported, Partially Supported, or Contradicted, and (iv) aggregates the outcomes into a single SA score via rubric-guided analysis. 
    Within the same reasoning trace, SG-PVR also evaluates video quality along three dimensions.}
    \label{figure:overview}
\end{figure*}

\section{Related Works}
\subsection{Reward Models for T2V Generation}

Non-reasoning RMs fine-tune a vision-language model (VLM) to emit scalar or per-dimension scores in a single pass~\cite{he2024videoscore, xu2026visionreward, liu2025improving, wang2024lift, wang2025unified}.
Reasoning RMs instead generate chain-of-thought rationales before scoring.
VideoScore2~\cite{he2025videoscore2} produces per-dimension rationales and pointwise scores, and UnifiedReward-Think~\cite{wang2025unifiedthink} reasons over multiple dimensions for pairwise ranking.
In all cases, however, semantic alignment is assessed without explicitly verifying each compositional requirement in the prompt.
SG-PVR instead decomposes the prompt into atomic verification claims and verifies each against the video and spatio-temporal scene graph evidence.

\subsection{Structured T2V Evaluation}

A separate line of work approaches T2V evaluation as training-free pipelines.
Holistic methods score prompt-video similarity in a shared embedding space~\cite{huang2024vbench, liu2024evalcrafter}, but report alignment as a single score without explicitly verifying individual prompt requirements, making failures difficult to localize.
%Holistic methods score prompt-video similarity in a shared embedding space~\cite{huang2024vbench, liu2024evalcrafter}, absorbing partial violations of compositional requirements into a single score. 

Decomposition-based approaches instead break each prompt into atomic elements to evaluate.
ETVA~\cite{guan2025etva} answers atomic yes/no questions about the prompt via multi-stage reasoning, and NeuS-V~\cite{sharan2025neuro} evaluates semantic alignment by computing the satisfaction probability of a video automaton against temporal logic specifications extracted from the prompt.
These provide more fine-grained evaluation than holistic similarity, but two limitations remain.

\emph{(i)} Their evaluation provides limited rationale for how each element is supported or violated by visual evidence. These methods are also limited in evaluating fine-grained temporal alignment, such as event sequences and state changes.
%\emph{(ii)} The final score aggregates judgments uniformly across heterogeneous semantic dimensions---entity, attribute, action, relation, and temporal order---so qualitatively different failures contribute identically.
\emph{(ii)} Final scores are obtained through uniform aggregation over decomposition outcomes, treating prompt elements equally regardless of their semantic importance.
SG-PVR addresses both within a single RM. 
It internalizes prompt decomposition into the model's reasoning trace, explicitly verifies atomic claims with raw-video and scene-graph evidence, and aggregates claim-level judgments through a rubric-guided final analysis rather than uniform averaging.
%Each claim is verified against a spatio-temporal scene graph that exposes supporting entities, relations, and time intervals as structured evidence.
%In addition, the verification plan organizes claims by semantic criterion, so the rubric-guided summary reflects the structure of semantic failures rather than averaging them uniformly.

\subsection{Scene Graph for Video Understanding}
Recent work uses video scene graphs as structured intermediates for video reasoning.
SG-VLM~\cite{ma2025bridging} couples a frozen VLM with query-aware scene graph filtering.
GraphThinker~\cite{cheng2026graphthinker} constructs event-level scene graphs as an intermediate thinking step. 
Video-of-Thought~\cite{fei2024video} decomposes video reasoning into sub-problems, grounded in spatio-temporal scene graphs.
These works share the premise that textualized scene graphs are effective grounding evidence for video reasoning, but target open-ended video understanding and consult the graph as a contextual signal during free-form reasoning. 
SG-PVR instead targets video reward modeling and pairs the graph with a verification plan, so the graph serves as the structured evidence for verifying each atomic claim.

\section{Method}
\label{sec:method}
\subsection{Overview}
SG-PVR is a video reward reasoning framework built on a single VLM that, given a text prompt $P$ and a video $V$, produces a reasoning trace and scores along two axes in a single generation: \emph{Semantic Alignment} (SA) and \emph{Quality}.
SA measures how well $V$ satisfies the requirements stated in $P$ through the SG-grounded plan-and-verify reasoning described in \S\ref{sec:sa}.
%SA measures faithfulness to the requirements stated in $P$ through SG-grounded plan-and-verify reasoning described in \S\ref{sec:sa}.
Quality assesses the intrinsic perceptual properties of $V$ as a separate axis from semantic alignment, through direct reasoning over $V$ under the predefined rubric described in \S\ref{sec:quality}.
%Quality assesses intrinsic perceptual properties of $V$ that are not specified by $P$; we evaluate it through direct reasoning over $V$ under a predefined rubric described in \S\ref{sec:quality}. 
Figure~\ref{figure:overview} illustrates the full reasoning steps.

\subsection{Semantic Alignment Evaluation: SG-Grounded Plan-and-Verify Reasoning}
\label{sec:sa}
SA evaluation runs in four steps. 
\textbf{Scene Graph Extraction} extracts a spatio-temporal scene graph (SG) $\mathcal{G}$ from $V$. 
\textbf{Verification Plan Generation} decomposes $P$ into atomic claims, each tagged Critical or Minor. 
\textbf{SG-Grounded Claim Verification} checks each claim against $V$ and $\mathcal{G}$. 
\textbf{Score Aggregation} consolidates the per-claim outcomes into a 1--5 SA score. 
% The full reasoning trace is enclosed in the \texttt{<semantic>} tag with sub-tags \texttt{<scene\_graph>}, \texttt{<plan>}, and \texttt{<analysis>}. 
Figure~\ref{fig:model_output} in Appendix~\ref{app:qualitativeexamples}  provides an example of this SA reasoning trace.

\paragraph{Spatio-temporal Scene Graph.} 
The scene graph captures the spatio-temporal content of $V$. 
It consists of an object scene graph, which represents entities and their attributes, and a relationship scene graph, which represents actions or spatial relations with time intervals. Together, they provide a structured representation of temporal relations in the video.
Formally, $\mathcal{G}$=($\mathcal{E}$, $\mathcal{R}$) consists of entities $\mathcal{E}$ with categories and attributes, and temporally grounded relations $\mathcal{R}$ of the form $(e_s, r, e_o, t)$ indicating that during time interval $t = [t_{\text{start}}, t_{\text{end}}]$ in seconds,  subject $e_s$ is related to object $e_o$ by predicate $r$. 
% Each entity $e \in \mathcal{E}$ is a tuple of a unique identifier, a category, and an attribute list; each relation in $\mathcal{R}$ is a directed tuple $(e_s, r, e_o, t)$ indicating that during the time interval $t = [t_{\text{start}}, t_{\text{end}}]$ (in seconds), the subject $e_s$ stands in predicate $r$ to the object $e_o$. 
$\mathcal{G}$ is serialized as JSON-like text and included in the VLM's reasoning context. 

\paragraph{Verification Plan Generation.} 
A Verification Claim is an atomic, individually verifiable statement asserting a single semantic requirement from $P$ (e.g., ``a cat is present'', ``the cat is black'').
The model produces a Verification Plan from $P$, a structured set of atomic claims specifying what should be checked in the generated video. 
We consider five semantic criteria: Entity, Attribute, Action, Spatial Relation, and Temporal Constraint. 
The model identifies which criteria are expressed by $P$, then decomposes $P$ into a minimal set of atomic Verification Claims under each identified criterion. 
Each claim states a single requirement to be evaluated from the video and its scene graph, with a Critical or Minor tag indicating its importance to the prompt. These tags are used to determine the final score, integrated with per-claim outcomes.
% Figure~\ref{fig:model_output} in Appendix~\ref{app:qualitativeexamples} illustrates this decomposition.

\paragraph{SG-Grounded Claim Verification.} 
Given the plan, the model judges each claim independently, using $V$ and $\mathcal{G}$ together as evidence. 
The two sources play complementary roles: $\mathcal{G}$ supports direct lookup for claims about entities, relations, and temporal structure, while $V$ supplies the visual evidence needed to recover entities or relations the graph may have missed and to ground claims that require pixel-level inspection.
% Figure~\ref{fig:model_output} in Appendix~\ref{app:qualitativeexamples} shows an example of claim-level judgments grounded in both the video and its scene graph.
For each claim, the model assigns one of three outcomes---\emph{Supported}, \emph{Partially Supported}, or \emph{Contradicted}---together with the supporting evidence drawn from $\mathcal{G}$, $V$, or both. 
The output is a structured claim-level analysis containing this evidence and the assigned outcome for every claim in the plan.

\paragraph{Score Aggregation.} 
Finally, the model assigns the SA score on a 1--5 scale using a predefined scoring rubric. The rubric defines score levels based on the distribution of Supported, Partially Supported, and Contradicted outcomes (e.g., score 5 corresponds to all claims Supported; score 1 to all Critical claims Contradicted). Before assigning the score, the model writes a short final analysis that integrates the judgments of Critical and Minor claims to evaluate how well the video semantically aligns with the prompt. The details of the rubric-based score aggregation are provided in Appendix~\ref{app:semantic_reasoning_trace_generation}.

\subsection{Quality Evaluation} 
\label{sec:quality}
The Quality axis assesses intrinsic perceptual properties of $V$ not specified by $P$. 
Since quality is prompt-independent, we evaluate it through holistic reasoning over $V$ along three perceptual dimensions, generated continuously after SA reasoning so a single generation produces both semantic and quality scores.
Visual Quality (VQ) focuses on per-frame visual fidelity, including resolution, blur, and artifacts. 
Temporal Quality (TQ) covers cross-frame coherence, including object identity persistence, flickering, and motion smoothness. 
Physical/Common-sense Consistency (PC) covers  real-world physics and common-sense conformity.
Each dimension is scored on a 1--5 scale under a rubric where 5 indicates virtually undetectable defects and 1 indicates frequent and disruptive defects.
% The three rationales and scores are emitted enclosed in the \texttt{<quality>} tag.

\subsection{Training Pipeline}
Training proceeds in two stages.
% Stage 1 teaches the model to extract spatio-temporal scene graphs, allowing Stage 2 to condition on a stably-formed graph rather than co-adapting to a still-learning graph generator.
Stage 1 first establishes scene-graph generation capability, avoiding the need to learn the intermediate representation from scratch jointly with reward reasoning. 
% Stage 2 then trains full reward reasoning behavior on top of the Stage 1 capability.
Stage 2 then jointly fine-tunes scene-graph generation and reward reasoning.

\subsubsection{Stage 1: Scene Graph Generation}
Stage 1 trains the base VLM to generate $\mathcal{G}$ from $V$. 
We construct training pairs $(V, \mathcal{G})$ from existing video scene graph datasets providing entities, attributes, and temporally localized relations annotated for each video.
A stronger VLM filters these raw annotations conditioned on the video and its accompanying prompt, retaining only the entities and relations central to the video's semantics while discarding incidental items and annotations the video clearly contradicts.
The base VLM is then fine-tuned on the resulting $(V, \mathcal{G})$ pairs with a standard next-token prediction objective. 
Full data construction details are provided in Appendix~\ref{sec:stage1_data}.

\subsubsection{Stage 2: Reasoning Supervised Fine-Tuning}
\label{sec:reasoning_sft}
Stage 2 fine-tunes the Stage 1 model to produce the evaluation trace in a single generation conditioned on a video and its prompt. 
% Each training sample is a video paired with one trace combining a \texttt{<semantic>} block in the plan-and-verify format and a \texttt{<quality>} block covering VQ, TQ, and PC, together with the SA score and the three quality scores. 
% The two blocks are constructed by separate procedures described below but jointly form the supervision target for a single video.
Each training sample is a video paired with one trace combining a \texttt{<scene\_graph>} block, a \texttt{<semantic>} block in the plan-and-verify format, and a \texttt{<quality>} block covering VQ, TQ, and PC, with the SA score and the three quality scores. 
The scene graph is generated using the Stage 1 checkpoint, whereas the semantic and quality blocks are constructed by separate procedures described below; these components form the supervision target for Stage 2.

\paragraph{SA Reasoning Data Construction.}
We apply the SA evaluation procedure of \S\ref{sec:sa} with a teacher VLM to generate multiple reasoning candidates per training sample, and retain only those passing three filters in order: format filtering for trace structure and plan adherence, exact matching to the ground-truth (GT) score provided by the source dataset, and consistency verification, which checks whether claim-level judgments are logically consistent and whether the final score follows the rubric. 
Format-failing candidates are refined within a small number of rounds; score-mismatched ones are regenerated with the GT score as the final-answer anchor; consistency-failing ones are refined with consistency feedback.
Providing the GT score during regeneration risks score-conditioned reasoning, which we limit by requiring per-claim outcomes to cite $V$ and $\mathcal{G}$ with no GT-aware information in the intermediate reasoning, and by rejecting traces whose cited evidence does not logically support the assigned judgments under the rubric.

\paragraph{Quality Reasoning Data Construction.}

We prompt the same teacher VLM to sample multiple candidate traces per video, each jointly producing rationales and 1--5 scores for VQ, TQ, and PC. 
Target scores are taken from the source dataset when available and otherwise from pseudo-labels generated by the scoring models described in Appendix~\ref{app:impdetails}. 
For each dimension, we select a rationale whose predicted score is within $\pm 1$ of the target and verify that its reasoning is consistent with the predicted score by masking the numerical score and re-inferring it from the rationale alone. 
For candidate rationales that pass this check, GPT-4o mini~\cite{openai2024gpt4omini} lightly adjusts the wording to match the target score, following VideoScore2~\cite{he2025videoscore2}, without introducing new visual observations. 
If no valid candidate rationale is available, we regenerate it conditioned on the target score under a prompt emphasizing visual grounding and retain it only if the same consistency check passes.

\paragraph{Training.}
We retain only videos where both SA and quality construction succeed, and combine the scene graph, SA reasoning trace, and quality reasoning trace into a single target sequence. 
The Stage 1 checkpoint is fine-tuned on this target with next-token prediction on response tokens, jointly updating scene graph generation, plan-and-verify reasoning, and quality reasoning with a single objective.

\section{Experiments}
\label{sec:experiments}
\subsection{Experimental Setup}

\begin{table*}[t!]
\centering
\resizebox{\textwidth}{!}{%
\begin{tabular}{@{}ll ccc cc c@{}}
\toprule
\multirow{2}{*}{\textbf{Method}} & \multirow{2}{*}{\textbf{Base Model}} & \multicolumn{3}{c}{\textbf{VideoScore-Bench-v2 (V/P/A)}} & \multicolumn{2}{c}{\textbf{LGVQ (S/T/A)}} & \textbf{MJ-Bench-Video} \\ \cmidrule(lr){3-5} \cmidrule(lr){6-7} \cmidrule(lr){8-8}
 &  & \textbf{Acc. (\%)} & \textbf{Relaxed Acc. (\%)} & \textbf{PLCC ($\times 100$)} & \textbf{SRCC ($\times 100$)} & \textbf{PLCC ($\times 100$)} & \textbf{Acc. (\%)} \\ \midrule
\multicolumn{8}{@{}l}{\textit{Metric-based Models}} \\
OmniScore* & - & - / - / 29.2 & - / - / 78.6 & - / - / 32.8 & - / - / 40.3 & - / - / 46.5 & 37.4 \\ \midrule
\multicolumn{8}{@{}l}{\textit{Non-reasoning Reward Models}} \\
VideoScore~v1.1$^{\dagger}$ & Mantis-8B & \textbf{41.5} / \textbf{38.9} / \textbf{34.9} & \textbf{91.0} / \textbf{87.0} / \underline{82.4} & \textbf{49.0} / \textbf{47.0} / 30.9 & \underline{30.5} / \textbf{39.1} / 19.6 & \underline{37.1} / \textbf{44.0} / \underline{21.4} & \textbf{71.6} \\
VisionReward* & CogVLM2-13B & - & - & - & - & - & 58.2 \\
VideoReward* & Qwen2-VL-2B & \underline{23.8} / - / 29.6 & \underline{61.0} / - / 67.4 & \underline{41.4} / - / \textbf{44.6} & \textbf{50.7} / - / \textbf{47.5} & \textbf{55.8} / - / \textbf{49.4} & 51.6 \\
LiFT* & VILA-13B & 14.2 / - / 13.2 & 48.4 / - / 47.8 & 10.2 / - / 31.2 & 26.3 / \underline{30.8} / \underline{20.8} & 23.4 / \underline{30.7} / 19.7 & 27.3 \\
UnifiedReward~2.0* & Qwen3.5-9B & - / \underline{32.3} / \underline{32.8} & - / \underline{84.6} / \textbf{84.0} & - / \underline{37.6} / \underline{35.8} & - / - / 9.4 & - / - / 14.5 & \underline{68.9} \\ \midrule
\multicolumn{8}{@{}l}{\textit{Reasoning-based Reward Models}} \\
VideoScore2* & Qwen2.5-VL-7B & \textbf{50.0} / \textbf{39.2} / \textbf{46.0} & \underline{92.4} / \textbf{87.4} / \textbf{90.4} & \textbf{62.9} / \underline{50.1} / \textbf{60.3} & \textbf{58.0} / - / \textbf{48.9} & \textbf{63.9} / - / \underline{49.9} & \underline{67.1} \\
VideoScore2 (SFT Only)* & Qwen2.5-VL-7B & 41.4 / 35.6 / 39.0 & 87.8 / 81.4 / \underline{83.8} & 48.2 / 42.4 / 53.0 & \underline{43.9} / - / 41.5 & \underline{52.1} / - / 43.2 & 64.0 \\ 
\textbf{SG-PVR} & Qwen3.5-9B & \underline{43.6} / \underline{38.2} / \underline{39.8} & \textbf{93.8} / \underline{83.2} / 82.6 & \underline{58.5} / \textbf{51.7} / \underline{56.6} & 42.3 / \textbf{51.7} / \underline{48.0} & 43.4 / \textbf{54.5} / \textbf{51.1} & \textbf{68.2} \\ \bottomrule
\end{tabular}%
}
\caption{
Performance comparison on reward model benchmarks. 
VideoScore-Bench-v2 reports V (Visual), P (Physical), A (Alignment); LGVQ reports S (Spatial), T (Temporal), A (Alignment). 
For all metrics, higher is better.
- indicates that the model does not provide a semantically corresponding native output for that dimension; we do not broadcast a single composite score across distinct dimensions.
$^{*}$ denotes results we evaluated using publicly available checkpoints, and $^{\dagger}$ denotes results reported in \citet{he2025videoscore2}.
The best score within each group is in \textbf{bold} and the second-best is \underline{underlined}.
}
\label{tab:reward_benchmarks}
\end{table*}

\begin{table}[t!]
\centering
\resizebox{\columnwidth}{!}{%
\begin{tabular}{@{}l ccc@{}}
\toprule
\textbf{Method} & \textbf{Task1 Acc.} & \textbf{Task2 Acc.} & \textbf{Overall (\%)} \\ \midrule
\multicolumn{4}{@{}l}{\textit{Training-free Baselines}} \\
OmniScore        & 47.7 & 53.8 & 6.0 \\
VBench-2.0        & 22.4 & 21.6 & 10.2 \\ 
ETVA             & 33.3 & 31.0 & 11.2 \\ \midrule
\multicolumn{4}{@{}l}{\textit{Non-reasoning Reward Models}} \\
VideoScore~v1.1    & 51.3 & 47.2 & 1.3 \\
VisionReward     & 20.2 & 19.0 & 0.5 \\
VideoReward      & 19.2 & 17.0 & 0.8 \\
LiFT             & 0.2  & 0.2  & 0.0 \\
UnifiedReward~2.0 & 58.8 & 52.5 & \textbf{30.8} \\ \midrule
\multicolumn{4}{@{}l}{\textit{Reasoning-based Reward Models}} \\
VideoScore2           & 42.0 & 40.4 & 16.1 \\
UnifiedReward-Flex    & 70.9 & 37.0 & 23.1 \\
UnifiedReward-Think   & 67.0 & 38.3 & 19.8 \\
\textbf{SG-PVR}    & 51.3 & 51.9 & \textbf{31.4} \\ \bottomrule
\end{tabular}%
}
\caption{
Performance on the temporal semantics test set. 
Task1 and Task2 measure accuracy under the positive and negative prompts respectively; 
Overall (joint correctness) requires both tasks correct on the same pair.
}
\label{tab:temporal_semantics}
\end{table}

\paragraph{Implementation Details.}
We build our model on Qwen3.5-9B~\cite{qwen3.5} as the student, distilled from Qwen3.5-27B as the teacher for data construction.
The student is trained in two supervised fine-tuning stages with standard next-token prediction.
Stage 1 learns scene graph generation on Synthetic Visual Genome 2 (SVG2)~\cite{gao2026synthetic}, resulting in approximately 307K pairs after filtering. 
Stage 2 fine-tunes the Stage 1 checkpoint on reasoning traces constructed from several pointwise-scored T2V preference datasets, yielding 83K samples after filtering. Full data construction and training details are in Appendix~\ref{app:impdetails}.

\paragraph{Benchmarks.}
We evaluate SG-PVR in three settings.
First, \textbf{pointwise reward prediction} measures whether a reward model's absolute scores align with human judgments on VideoScore-Bench-v2 (VSB-v2)~\cite{he2025videoscore2}, LGVQ~\cite{zhang2025benchmarking}, and MJ-Bench-Video~\cite{tong2025mj} in terms of accuracy and correlation. 

Second, \textbf{fine-grained temporal semantic understanding} tests whether an evaluator distinguishes videos differing only in temporal structure.
We construct a 600-pair test set from filtered Vinoground~\cite{zhang2024vinoground} and TimeBlind~\cite{li2026timeblind}, where each pair consists of two videos and two prompts differing only in temporal semantics.
Task1 is correct when the positive video scores higher than the negative under the positive prompt; Task2 applies the symmetric criterion under the negative prompt.
Overall measures joint correctness, requiring both Task1 and Task2 to be correct for the same instance. 
The full construction protocol is provided in Appendix~\ref{app:temporal_details}, and we will release the evaluation code and benchmark construction metadata.
% Third, \textbf{test-time reranking} evaluates downstream utility on T2V-CompBench~\cite{sun2025t2v}. 
% Using Wan2.1-1.3B~\cite{wan2025wan}, we sample 8 candidate videos per prompt and rerank each pool with each reward model, taking the top-ranked candidate as the final output. 
% Full details are in Appendix~\ref{app:benchmarks}.

Third, \textbf{test-time reranking} evaluates downstream utility on T2V-CompBench~\cite{sun2025t2v}.
Using Wan2.1-1.3B~\cite{wan2025wan} and CogVideoX-2B~\cite{yang2024cogvideox}, we sample 8 candidates per prompt from each generator and select the highest-scoring candidate under each reward model.
Benchmark details are provided in Appendix~\ref{app:benchmarks}.

\paragraph{Baselines.}
We compare SG-PVR against three families of pointwise-scoring baselines on reward model benchmarks: metric-aggregation methods~\cite{liu2025videodpo}, non-reasoning reward models~\cite{he2024videoscore,xu2026visionreward,liu2025improving,wang2024lift,wang2025unified}, and reasoning-based reward models~\cite{he2025videoscore2}. 
For the temporal semantics test set, we additionally include pairwise-only RMs~\cite{wang2025unifiedthink, wang2026unified}, the decomposition-based ETVA~\cite{guan2025etva}, and a VLM-prompting baseline following VBench-2.0~\cite{zheng2025vbench}.
For test-time reranking, we compare against OmniScore~\cite{liu2025videodpo}, VideoScore2~\cite{he2025videoscore2}, and random selection. 
For SG-PVR, we average the SA score with the mean of VQ, TQ, and PC.
For VideoScore2, we average its three output dimensions, and for OmniScore we
use its scalar reward directly.
Further details are provided in Appendix~\ref{app:baselines}.

\subsection{Results}
We organize our analysis by the three evaluation settings.
Since SG-PVR is trained with SFT only, we primarily compare it against SFT-only reasoning-based RMs; RL-trained models such as VideoScore2 serve as a separate reference.

\paragraph{Pointwise Reward Prediction.}
Table~\ref{tab:reward_benchmarks} reports pointwise scoring on VSB-v2, LGVQ, and MJ-Bench-Video. 
SG-PVR achieves higher semantic alignment performance than most metric-based and non-reasoning RMs on VSB-v2 and LGVQ, and records the highest MJ-Bench-Video accuracy among reasoning-based RMs.
Compared with VideoScore2 (SFT Only), SG-PVR achieves higher Alignment Accuracy and PLCC on VSB-v2 and higher Alignment PLCC on LGVQ.
% The modest VSB-v2 Alignment Accuracy gap (39.8\% vs.\ 39.0\%) reflects both VideoScore2's in-domain advantage and opposite trends across event complexity. 
Despite VideoScore2’s in-domain advantage, SG-PVR achieves a modest gain in VSB-v2 Alignment Accuracy (39.8\% vs. 39.0\%).
VSB-v2 is derived from VideoFeedback2, the training dataset of VideoScore2, while VideoFeedback2 accounts for only 21\% of SG-PVR’s Stage 2 data.
Moreover, SG-PVR underperforms on single-event prompts (40.52\% vs.\ 45.69\%) but substantially outperforms VideoScore2 on prompts with two or more events (41.01\% vs.\ 34.53\%), indicating a clearer advantage on prompts involving multiple events.

We further control for backbone strength by reproducing the VideoScore2 SFT-only recipe on the Qwen3.5-9B backbone.
This controlled model reaches 63.6 on MJ-Bench-Video and 11.9 Temporal Overall, below SG-PVR's 68.2 and 31.4, showing that backbone strength alone does not explain the gains, as detailed in Appendix~\ref{app:base_model}.
SG-PVR also achieves the highest LGVQ Temporal scores among models that provide a Temporal output.
On perceptual quality, it outperforms VideoScore2 (SFT Only) across the VSB-v2 Visual and Physical metrics but trails on LGVQ Spatial.
Against the RL-trained VideoScore2, SG-PVR leads on VSB-v2 Physical PLCC and LGVQ Alignment PLCC, while trailing on the remaining pointwise metrics.

\paragraph{Fine-grained Temporal Semantic Understanding.}
Table~\ref{tab:temporal_semantics} evaluates fine-grained temporal semantics using prompt-video quadruples where the two videos differ only in temporal structure.
SG-PVR achieves the highest Overall accuracy among all compared methods.
UnifiedReward-Flex and UnifiedReward-Think are evaluated in pairwise mode with the same video pair under both the positive and negative prompts.
% Their high Task1 accuracy but much lower Task2 accuracy suggests a strong input-order bias: the selected video is more affected by its input position than by the prompt semantics.
Their high Task1 accuracy but much lower Task2 accuracy is consistent with a strong input-order bias under the fixed presentation order: the models tend to favor the first video even when the prompt semantics switch to favor the second.
In contrast, SG-PVR scores each prompt-video pair pointwise, avoiding this pairwise input-order issue and maintaining balanced Task1 and Task2 accuracies with only a 0.6-point gap.
% ETVA, despite using prompt decomposition, obtains lower Overall accuracy than SG-PVR.
% Its evaluation does not provide structured visual evidence for each atomic check, and its uniform aggregation treats all checks equally regardless of their semantic importance.
% This contrasts with SG-PVR, which verifies each claim with evidences and aggregates claim judgments through rubric-guided analysis.
ETVA also uses prompt decomposition but obtains lower Overall accuracy than SG-PVR. 
Its evaluation lacks structured visual evidence for each atomic check and uniformly aggregates the outcomes regardless of their semantic importance. 
In contrast, SG-PVR grounds each claim in explicit evidence and uses rubric-guided aggregation.

% \begin{table}[t!]
% \centering
% \resizebox{\columnwidth}{!}{%
% \begin{tabular}{@{}l ccccccc@{}}
% \toprule
% \textbf{Method} & \textbf{C-Attr} & \textbf{D-Attr} & \textbf{Spat.} & \textbf{Mot.} & \textbf{Act.} & \textbf{Inter.} & \textbf{Num.} \\
% \midrule
% Random          & 77.5 & \underline{2.7}  & 53.4 & 38.3 & 61.9 & 59.5 & 45.7 \\
% OmniScore       & \underline{80.7} & 2.1  & \underline{55.3} & \textbf{39.5} & \underline{67.6} & \textbf{65.3} & 49.6 \\
% VideoScore2     & 76.4 & 1.4  & 50.6 & 38.5 & 62.8 & 59.2 & \underline{49.8} \\
% \midrule
% \textbf{SG-PVR} & \textbf{81.7} & \textbf{4.2} & \textbf{58.7} & \underline{39.3} & \textbf{69.1} & \underline{63.6} & \textbf{58.8} \\
% \bottomrule
% \end{tabular}%
% }
% \caption{Test-time reranking on T2V-CompBench: per-dimension score of the top-ranked candidate.
% For all dimensions, higher is better.
% Dimensions: C-Attr (Consistent Attribute), D-Attr (Dynamic Attribute), Spat. (Spatial Relationship), Mot. (Motion), Act. (Action), Inter. (Interaction), Num. (Numeracy).}
% \label{tab:reranking}
% \end{table}

\begin{table}[t!]
\centering
\resizebox{\columnwidth}{!}{%
\begin{tabular}{@{}llccccccc@{}}
\toprule
\textbf{Generator} & \textbf{Method} & \textbf{C-Attr} & \textbf{D-Attr} & \textbf{Spat.} & \textbf{Mot.} & \textbf{Act.} & \textbf{Inter.} & \textbf{Num.} \\ \midrule
Wan2.1-1.3B  & Random          & 77.5          & 2.7          & 53.4          & 38.3          & 61.9          & 59.5          & 45.7          \\
             & OmniScore       & {\ul 80.7}    & 2.1          & 55.3          & \textbf{39.5} & {\ul 67.6}    & \textbf{65.3} & 49.6          \\
             & VideoScore2     & 78.3          & 3.1          & 55.9          & 38.8          & 67.4          & 60.6          & 55.3          \\ \cmidrule(l){2-9} 
             & \textbf{SG-PVR} & \textbf{81.7} & \textbf{4.2} & \textbf{58.7} & {\ul 39.3}    & \textbf{69.1} & {\ul 63.6}    & \textbf{58.8} \\ \midrule
CogVideoX-2B & Random          & 77.9          & 1.4          & 50.0          & 38.3          & 48.4          & 59.2          & 36.4          \\
             & OmniScore       & 80.2          & 1.0          & {\ul 53.1}    & \textbf{39.3} & {\ul 56.5}    & \textbf{62.1} & 37.4          \\
             & VideoScore2     & {\ul 82.2}    & {\ul 1.5}    & 52.4          & 38.8          & 55.6          & 60.4          & {\ul 42.9}    \\ \cmidrule(l){2-9} 
             & \textbf{SG-PVR} & \textbf{83.6} & \textbf{2.5} & \textbf{59.6} & {\ul 39.0}    & \textbf{60.5} & {\ul 62.0}    & \textbf{47.1} \\ \bottomrule
\end{tabular}
}
\caption{Test-time reranking on T2V-CompBench: per-dimension score of the top-ranked candidate.
For all dimensions, higher is better.
Dimensions: C-Attr (Consistent Attribute), D-Attr (Dynamic Attribute), Spat. (Spatial Relationship), Mot. (Motion), Act. (Action), Inter. (Interaction), Num. (Numeracy).}
\label{tab:reranking}
\end{table}

\paragraph{Test-time Reranking.}
% Table~\ref{tab:reranking} reports test-time reranking on T2V-CompBench, where we rerank $N{=}8$ candidates from Wan2.1, scoring each dimension with T2V-CompBench's per-dimension evaluator.
% SG-PVR outperforms Random and VideoScore2 on every dimension, and leads on five of seven against the stronger OmniScore baseline.
% The largest gains concentrate on Numeracy, Spatial Relations, Action, and Consistent Attribute, dimensions that plan-and-verify explicitly verifies per claim, confirming fine-grained alignment scoring transfers to downstream generation.

Table~\ref{tab:reranking} reports test-time reranking on T2V-CompBench using $N{=}8$ candidate videos from each of Wan2.1-1.3B and CogVideoX-2B.
Across both generators, SG-PVR outperforms Random and VideoScore2 on all seven dimensions and leads OmniScore on five of seven dimensions.
The gains are particularly pronounced on Spatial Relationship, Action, and Numeracy, which require fine-grained verification of compositional prompt requirements.
% The consistent improvements across two different generators show that SG-PVR's alignment scoring transfers to downstream candidate selection beyond a single T2V generator.
The consistent improvements across both generators show that SG-PVR’s alignment scoring transfers beyond a single T2V generator. 
These gains also persist with semantic-alignment-only reranking, as shown in Appendix~\ref{app:semantic_reranking}.
% We further verify that these gains persist when reranking with semantic-alignment scores alone in Appendix~\ref{app:semantic_reranking}.

\begin{table}[t!]
\centering
\resizebox{\columnwidth}{!}{%
\begin{tabular}{@{}l cc cc@{}}
\toprule
\multirow{2}{*}{\textbf{Method}} & \multicolumn{2}{c}{\textbf{Setting}} & \textbf{VSB-v2} & \textbf{Temp.} \\
\cmidrule(lr){2-3}
 & \makecell{\textbf{Grounding}\\\textbf{Information}} & \makecell{\textbf{Reasoning}\\\textbf{Structure}} & \textbf{Acc (V/P/A)} & \textbf{Overall} \\
\midrule
\textbf{SG-PVR} & V+SG & Plan-and-Verify & \textbf{43.6} / \textbf{38.2} / 39.8 & \textbf{31.4} \\
\midrule
w/o SG & V & Plan-and-Verify & 42.8 / 31.8 / 39.2 & \underline{30.5} \\
w/ Caption & V+Caption & Plan-and-Verify & \underline{43.4 }/ 33.6 / 36.0 & 26.2 \\
w/o Explicit Plan & V+SG & Direct Verification & 41.0 / 34.0 / 38.4 & 28.0 \\
w/o P\&V & V+SG & Free-form & 42.0 / \underline{36.8} / \underline{42.8} & 21.8 \\
w/o SG and P\&V & V & Free-form & 39.8 / 35.2 / 35.6 & 25.6 \\
w/o Reasoning & V & --- & 41.8 / 34.0 / \textbf{45.4} & 20.2 \\
\bottomrule
\end{tabular}%
}
\caption{Ablation on key design choices. 
We report VSB-v2 Accuracy (V/P/A) and joint correctness (Overall) on the temporal semantics test set.}
\label{tab:ablation}
\end{table}

\paragraph{Statistical Robustness.}
Across three independent training seeds, including the run reported in the main table, SG-PVR achieves 29.7$\pm$1.7 on Temporal Overall and 67.6$\pm$1.3 on MJ-Bench-Video Overall, where $\pm$ denotes the standard deviation across runs, indicating stable performance across runs.
We further perform paired bootstrap testing with 10,000 resamples against our re-evaluation of VideoScore2 (SFT Only) under the same evaluation protocol.
The 95\% confidence intervals for the improvements are [15.9, 24.7] percentage points on Temporal Overall and [2.0, 6.2] points on MJ-Bench-Video Overall. 
Both intervals exclude zero, indicating statistically significant improvements over the SFT-only baseline.

\subsection{Ablation Studies}
% We conduct ablation studies along two orthogonal axes, as shown in Table~\ref{tab:ablation}: the evidence source for claim verification and the reasoning structure.
We conduct ablations along two axes shown in Table~\ref{tab:ablation}, varying the evidence source for claim verification and reasoning structure.
% For each setting, we construct the supervision targets to match the corresponding grounding information and reasoning format, and independently fine-tune the same Qwen3.5-9B backbone on the same Stage-2 source data.
For each setting, we construct the corresponding supervision from the same Stage-2 source data and independently fine-tune the same Qwen3.5-9B backbone.
% Evidence ablations compare three grounding conditions: video and scene graph, video-only, and video and unstructured caption. 
Evidence ablations compare video with scene graph, video only, and video with an unstructured caption.
% Structure ablations vary the reasoning format across three levels: free-form reasoning, direct verification without an explicit plan, and the full plan-and-verify trace.
% Reasoning ablations vary the supervision format across four levels: direct scoring without intermediate reasoning, free-form reasoning, direct verification without an explicit plan, and the full plan-and-verify trace.
Reasoning ablations compare direct scoring without intermediate reasoning, free-form reasoning, direct verification without an explicit plan, and full plan-and-verify reasoning.

\paragraph{Effect of SG-Grounded Evidence.}
We isolate the role of the scene graph while keeping the plan-and-verify structure fixed. 
The w/o SG variant verifies each claim using only the raw video, while w/ Caption replaces the scene graph with a detailed free-form caption. 
Both variants show lower VSB-v2 Accuracy and Temporal Overall than SG-PVR. 
% These results indicate that the structured scene graph provides useful grounding beyond the raw video and is more effective than unstructured textual context for representing temporally grounded visual evidence during claim verification.
These results show that the structured scene graph provides useful grounding beyond the raw video and represents temporally grounded evidence more effectively than unstructured textual context.

\paragraph{Effect of Reasoning Structure.}
% We isolate the role of structured reasoning while keeping the evidence fixed at $V$ and $\mathcal{G}$. 
% The w/o Explicit Plan variant removes the explicit plan-output stage but retains per-claim verification: the model implicitly determines the claims to verify during reasoning, without first externalizing them as a separate verification plan.
% The w/o Plan-and-Verify variant removes per-claim verification and instead generates a free-form rationale before scoring. 
% Removing the explicit plan decreases both VSB-v2 accuracy and Temporal Overall, suggesting that specifying the verification targets before claim verification improves both standard pointwise scoring and temporal semantic evaluation.
% The w/o Plan-and-Verify variant achieves higher VSB-v2 Alignment Accuracy than SG-PVR, but its Temporal Overall drops substantially from 31.4 to 21.8.
% This suggests that free-form reasoning can still produce reasonable scores on standard pointwise alignment benchmarks, but is less reliable when the evaluation requires consistent temporal semantic judgments.
% Removing both the scene graph and plan-and-verify structure further lowers VSB-v2 performance and remains below SG-PVR on Temporal Overall, suggesting that structured visual evidence and structured reasoning provide complementary gains.
% We further analyze reasoning trace faithfulness in Appendix~\ref{app:trace_analysis}.

We analyze the effect of reasoning structure with the grounding information fixed at $V+\mathcal{G}$.
The w/o Explicit Plan variant removes the separate plan-generation step, so the model identifies prompt requirements during verification and evaluates each individually rather than specifying them in advance.
The w/o Plan-and-Verify variant further removes claim-level verification and uses free-form reasoning before scoring.
Temporal Overall decreases from 31.4 to 28.0 without the explicit plan and to 21.8 with free-form reasoning.
We further compare reasoning structures under video-only grounding.
The w/o Reasoning variant directly predicts the final scores without an intermediate reasoning trace, while w/o SG and P\&V uses free-form reasoning and w/o SG uses the full plan-and-verify structure.
Temporal Overall increases from 20.2 without reasoning to 25.6 with free-form reasoning and 30.5 with plan-and-verify.
These results show that progressively structuring intermediate reasoning improves the model's ability to verify fine-grained spatio-temporal semantics.

Despite this advantage, w/o P\&V achieves higher VSB-v2 alignment accuracy. 
To understand this discrepancy, we stratify all 500 VSB-v2 examples by temporal event count and the number of claims generated by SG-PVR, using the median claim count of eight as the split. 
The largest degradation occurs for zero-event prompts with more than eight claims (35.71\% vs.\ 53.57\%), which predominantly describe static scene properties. 
This suggests that over-decomposing static descriptions into many atomic claims can introduce verification and aggregation errors. 
In contrast, the two methods perform identically on single-event prompts, while SG-PVR slightly outperforms w/o P\&V on prompts with at least two events when more than eight claims are generated (43.83\% vs.\ 43.21\%). 
These results suggest that the aggregate pointwise gap is largely driven by over-decomposition of static scene descriptions, whereas structured claim verification becomes more useful when multi-event prompts require many fine-grained conditions to be verified separately.
We further analyze reasoning trace faithfulness in Appendix~\ref{app:trace_analysis}.

\section{Analysis}
\label{sec:Analysis}
% \subsection{Fine-Grained Analysis by Temporal Event Complexity and Prompt Type}
\subsection{Fine-Grained Analysis}
\label{sec:finegrained_analysis}

\begin{figure}[t!]
\centering
\includegraphics[width=\columnwidth]{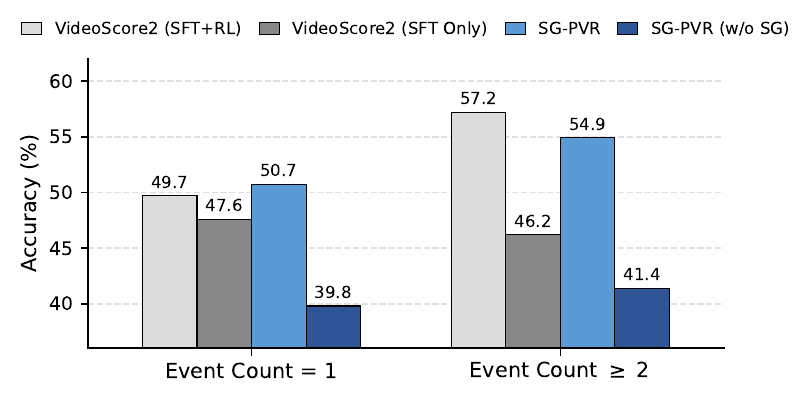}
\caption{
Pointwise alignment accuracy by event count on MJ-Bench-Video. 
%SG-PVR outperforms VideoScore2 (SFT Only) and SG-PVR (w/o SG) for both single-event and multi-event prompts, while remaining competitive with VideoScore2 (SFT+RL).
}
\label{fig:mjbench_4model_compare}
\end{figure}

\paragraph{Temporal Event Complexity Analysis.}
To analyze how alignment accuracy changes with temporal event complexity, we use GPT-5.4~\cite{openai2026gpt54} to decompose each MJ-Bench-Video prompt into temporally distinct visual events based on temporal markers (e.g., before, after, then) and compute its event count.
As shown in Figure~\ref{fig:mjbench_4model_compare}, we compare pointwise alignment accuracy between single-event and multi-event prompts. 

For single-event prompts, SG-PVR achieves the highest alignment accuracy among all baselines. 
% For multi-event prompts, SG-PVR reaches 54.9\%, outperforming both VideoScore2 (SFT Only) and SG-PVR (w/o SG). Notably, despite being trained only with SFT, SG-PVR remains competitive with VideoScore2 (SFT+RL) with a slight gap. 
% Overall, SG-PVR improves alignment accuracy over the SFT-only baseline for prompts involving temporal events and remains close to the RL-trained model, while the gain over SG-PVR (w/o SG) shows that scene graph grounding helps when multiple temporally connected events must be verified.
For multi-event prompts, SG-PVR reaches 54.9\%, outperforming VideoScore2 (SFT Only) and SG-PVR (w/o SG) while remaining close to VideoScore2 (SFT+RL). 
The gain over SG-PVR (w/o SG) indicates that scene graph grounding helps when multiple temporally connected events must be verified.

\begin{table}[t!]
\centering
\resizebox{\columnwidth}{!}{%
\begin{tabular}{@{}l cc cc@{}}
\toprule
    \textbf{Prompt Type} & \makecell{\textbf{VideoScore2}\\\textbf{(SFT+RL)}} & \makecell{\textbf{VideoScore2}\\\textbf{(SFT Only)}} & \textbf{SG-PVR} & \textbf{w/o SG} \\
    \midrule
    Entity Interaction  & \textbf{53.7} & 44.5 & 46.3 & 31.6 \\
    State Transition    & 70.0 & 70.0 & \textbf{71.2} & 53.8 \\
    Static Composition  & 44.4 & 40.4 & \textbf{46.4} & 37.5 \\
    Action Dynamics     & 52.1 & 45.4 & \textbf{55.0} & 42.6 \\
    Other               & \textbf{57.8} & 48.1 & 50.5 & 35.0 \\
    \bottomrule
\end{tabular}%
}
\caption{
Pointwise alignment accuracy by prompt type on MJ-Bench-Video. 
%SG-PVR achieves the best accuracy on State Transition, Static Composition, and Action Dynamics, and consistently improves over SG-PVR (w/o SG) across all prompt types.
}
\label{tab:mjbench_prompt_type}
\end{table}

\paragraph{Prompt Type Analysis.}
To examine whether the model is effective across different types of semantic requirements, we further analyze alignment accuracy by prompt type, as shown in Table~\ref{tab:mjbench_prompt_type}.
Using GPT-5.4, we classify each MJ-Bench-Video prompt into one of five categories based on its core semantic focus: Entity Interaction, State Transition, Static Composition, Action Dynamics, and Other. 
These categories distinguish prompts by whether they primarily involve entity/object interactions, visible state changes, non-interactive actions or motions, or static visual composition such as object presence, attributes, scene setting, and layout.

SG-PVR achieves the best accuracy on State Transition, Static Composition, and Action Dynamics, outperforming all baselines. 
Compared with SG-PVR (w/o SG), consistent gains across all prompt types confirm the contribution of scene graph grounding to fine-grained semantic verification. 
% Against VideoScore2 baselines, SG-PVR surpasses the SFT-only model across all prompt types and the RL-trained model on three categories, indicating the proposed reasoning structure achieves strong fine-grained alignment without RL training.
% Against VideoScore2 baselines, SG-PVR surpasses the SFT-only model across all prompt types and the RL-trained model on three categories, showing strong fine-grained alignment under SFT-only training.
SG-PVR surpasses VideoScore2 (SFT Only) across all prompt types and the RL-trained model on three categories, showing strong fine-grained alignment under SFT-only training.

\subsection{Scene Graph Analysis}
\label{sec:sg_analysis}
% We evaluate $\mathcal{G}$ along two properties: faithfulness (whether entries are grounded in the video) and coverage (whether the graph captures prompt-relevant content enough), assessed on 220 prompt-video pairs from DREAM-1k~\cite{wang2024tarsier} using GPT-5.4. 
% For faithfulness, entity precision reaches 98.9\%, attribute precision 89.2\%, and relation precision 81.3\%, indicating that $\mathcal{G}$ entries are reliably grounded in the video, with the moderate drop on relations reflecting the inherent difficulty of extracting fine-grained interactions. 
% For coverage, requirements are met at 23.4\% on average (43.6\% entities, 27.0\% attributes, 11.5\% actions), reflecting the limited expressiveness of the scene graph format for fine-grained events. 
% This gap is mitigated by our joint $V+\mathcal{G}$ design, where requirements not covered by $\mathcal{G}$ are recovered through raw video verification. 
% We examine the empirical contribution of this moderate coverage in the following section. 

We evaluate $\mathcal{G}$ along faithfulness and coverage on 220 prompt-video pairs from DREAM-1k~\cite{wang2024tarsier} using GPT-5.4.
Faithfulness measures whether the entities, attributes, and relations in $\mathcal{G}$ are actually supported by the video, while coverage measures how much prompt-relevant content depicted in the video is captured by $\mathcal{G}$. 
We compute coverage by decomposing each prompt into atomic requirements, excluding those absent from the video, and measuring the fraction represented in $\mathcal{G}$.
Entity, attribute, and relation precision reach 98.9\%, 89.2\%, and 81.3\%, respectively, indicating that the information in $\mathcal{G}$ is generally reliable, while the lower relation precision reflects the difficulty of extracting fine-grained interactions.

Overall requirement coverage is 23.4\%, with coverage rates of 43.6\% for entities, 27.0\% for attributes, and 11.5\% for actions.
The moderate coverage reflects the role of $\mathcal{G}$ as a compact, high-precision representation of salient entities and temporally grounded relations rather than an exhaustive description of the video.
Since claim verification uses both $V$ and  $\mathcal{G}$, requirements absent from $\mathcal{G}$ can still be evaluated from the raw video.
% Nevertheless, the w/o SG ablation consistently degrades temporal performance under the same plan-and-verify procedure, showing that $\mathcal{G}$ contributes complementary evidence beyond the reasoning structure, particularly for temporally complex prompts.
Nevertheless, the w/o SG ablation consistently degrades temporal performance under the same plan-and-verify procedure, with a pronounced effect on multi-event MJ-Bench-Video prompts.

\begin{figure}[t!]
\centering
\includegraphics[width=\columnwidth]{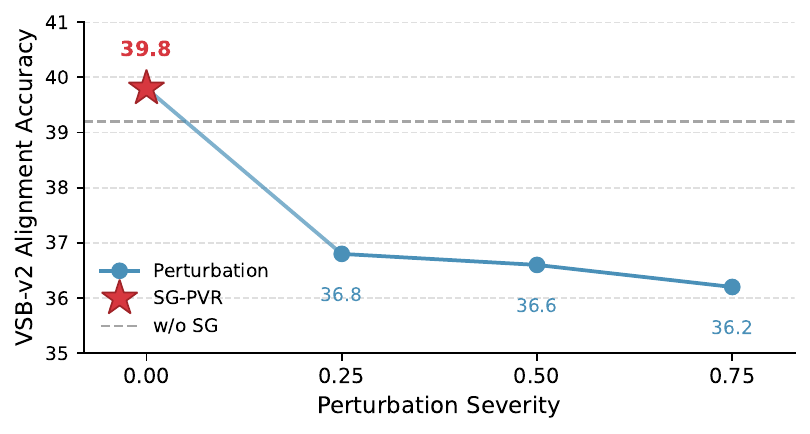}
\caption{Effect of scene graph perturbation on VSB-v2 Alignment Accuracy.}
\label{fig:sg_perturbation}
\end{figure}

% Beyond characterizing $\mathcal{G}$'s quality, we probe whether the RM actively consults $\mathcal{G}$ during reasoning. 
% We perturb it at inference time by dropping and replacing entries at severity levels \{0.25, 0.5, 0.75\} and measuring VSB-v2 Alignment Accuracy. 
% Any perturbation incurs up to 3-point drop, but additional corruption beyond 25\% does not compound it, suggesting the RM detects deviation from the self-extracted $\mathcal{G}$ and partially falls back on the raw video. 
% Perturbed configurations fall below the w/o SG baseline, though this reflects distributional shift rather than corrupted $\mathcal{G}$ being inherently worse than no $\mathcal{G}$. 
% Together, these results confirm that the RM actively consults $\mathcal{G}$.

Beyond evaluating $\mathcal{G}$ itself, we probe whether the RM actively consults $\mathcal{G}$ during reasoning by dropping and replacing entries at severity levels \{0.25, 0.5, 0.75\}.
As shown in Figure~\ref{fig:sg_perturbation}, perturbation reduces VSB-v2 Alignment Accuracy by up to 3.6 points, with little further degradation beyond 25\% corruption.
To explain this saturation, we track claims associated with corrupted entries.
Deleting claim-related SG entries changes only 11.2\% and 11.3\% of the corresponding judgments at severity 0.25 and 0.5, while inserting false entries changes about 6--7\% across perturbation levels.
Among unchanged judgments, explicit SG citation falls from 46\% without perturbation to 23\% and 17\% at severity 0.25 and 0.5, indicating greater reliance on raw-video evidence as the graph becomes less reliable.
The bounded degradation reflects fallback to raw video rather than failure to use $\mathcal{G}$. 
Combined with the w/o SG ablation, results show that $\mathcal{G}$ provides complementary evidence when reliable, while raw video limits error propagation under corruption.

\subsection{Computational Cost}

\begin{table}[t!]
\centering
\resizebox{\columnwidth}{!}{
\begin{tabular}{@{}lccc@{}}
\toprule
\textbf{Model} & \textbf{\begin{tabular}[c]{@{}c@{}}Input / Output\\tokens (mean)\end{tabular}} & \textbf{\begin{tabular}[c]{@{}c@{}}Wall-clock / video\\(s)\end{tabular}}  & \textbf{\begin{tabular}[c]{@{}c@{}}Throughput\\(videos/hr)\end{tabular}} \\ \midrule
VideoScore2 (7B) & 3132 / 418  & 0.77 & 4692 \\
SG-PVR (9B)      & 3001 / 1087 & 1.35 & 2675 \\
- w/o SG         & 2751 / 964  & 1.18 & 3056 \\
- w/o P\&V        & 2740 / 724  & 0.94 & 3823 \\
- w/o SG and P\&V   & 2452 / 568  & 0.94 & 3811 \\ \bottomrule
\end{tabular}
}
\caption{Inference cost comparison on 200 samples of VSB-v2. 
All models use the same samples and serving setup.}
\label{tab:inference_cost}
\end{table}

We measure inference cost under identical serving conditions using vLLM~\cite{kwon2023efficient} on a single A100 80GB GPU over 200 VSB-v2 samples, with 16 uniformly sampled frames per video.
SG-PVR requires 1.35 s/video compared with 0.77 s/video for VideoScore2, corresponding to a 1.8× overhead, and generates 1,087 rather than 418 output tokens on average. 
Ablations show that structured reasoning contributes more inference overhead than scene-graph extraction. 
SG-PVR still processes approximately 2,700 videos per hour, remaining practical for test-time reranking and post-training.
Additional analyses and qualitative examples are provided in Appendices~\ref{app:additional_analysis} and~\ref{app:qualitativeexamples}.

\section{Conclusion}
\label{sec:conclusion}
We present SG-PVR, a video reward model that performs semantic alignment evaluation through plan-and-verify reasoning grounded in a spatio-temporal scene graph.
The prompt is decomposed into atomic claims, which are verified against graph and video evidence before being aggregated through rubric-guided summarization.
The same reasoning trace also produces visual, temporal, and physical/common-sense quality scores.
Across the three evaluation settings, SG-PVR achieves strong performance on compositional and temporal semantic alignment, yielding a reward signal useful for both evaluation and downstream T2V generation.

\section*{Limitations}
The Stage 1 scene-graph extractor provides only partial coverage, particularly for action-level requirements. 
Joint verification with the raw video mitigates missing graph evidence, but improving temporal-event coverage remains important. 
Our evaluation includes videos up to 55 seconds, but does not establish robustness on minute-scale or longer-form videos, where scene-graph extraction and verification may become more challenging as temporal complexity grows. 
SG-PVR also evaluates Physical/Common-sense Consistency as a prompt-independent intrinsic property, which may penalize videos that intentionally depict fantastical or stylized behavior; a prompt-conditioned formulation may better accommodate such cases. 
Finally, SG-PVR is trained with SFT only, without reinforcement learning, due to the additional computational cost of RL for long-context video reasoning. 
Integrating RL with the proposed structured reasoning framework remains future work.

\section*{Acknowledgments}
This work was supported by the National Research Foundation of Korea (NRF) grant funded by the Korean government (MSIT) in 2026(Grant No. RS-2026-25613293, Project Title: Physics- and Mathematics-Informed Causal AI Models for Robust Analysis and Prediction under Geometry, Boundary-Condition, Parameter, and Multi-Environment Shifts, 20\%) and (Grant No. RS-2026-25473475, Project Title: Functional Analytic Approach to Foundational Theory of Stochastic PDE based Probabilistic Generative Models, 20\%), and was supported by the Institute of Information \& Communications Technology Planning \& Evaluation(IITP) grant funded by the Korea government(MSIT) (IITP-2026-RS-2024-00436857, ITRC(Information Technology Research Center), 10\%), (No. RS-2026-25507543, Development of AI Co-Scientist based on Scientific Causal World Model, 20\%), (No. RS-2022-II220612, Project Title: Geometric and Physical Commonsense Reasoning based Behavior Intelligence for Embodied AI, 20\%), and (No.RS-2026-25551943, Operation of Military-Industry-Academy Cooperation Center(Seoul(Yongsan)), 10\%), and was also supported by the AI Computing Infrastructure Enhancement (GPU Rental Support) User Support Program (RQT-25-090166) funded by the Ministry of Science and ICT (MSIT), Republic of Korea.

\bibliography{custom}

\appendix

\section{Broader Impacts}
\label{app:broaderimpacts}
SG-PVR is a video reward model for text-to-video generation. 
More faithful and interpretable reward signals can benefit creative media production, accessibility tools, and educational content synthesis, while contributing to more transparent evaluation and alignment practices. 
However, improvements in T2V reward modeling can also enhance the capabilities of video generation systems, which may be misused to produce deceptive content such as deepfakes. 
Additionally, preference datasets used for training may carry subjective biases that, if amplified during alignment, could lead generators to systematically favor certain visual or cultural representations. 
Our training and evaluation data are drawn entirely from publicly released datasets and benchmarks. 
The majority of videos are AI-generated synthetic content produced by text-to-video models, while natural videos also appear in the SA-V subset of SVG2 and in temporal-reasoning benchmarks such as Vinoground.
We rely on the data curation and release procedures of the original datasets and do not perform additional personally identifying information (PII) or offensive-content screening. 
We acknowledge this as a limitation, particularly for downstream uses that may require stricter content auditing.

\section{Implementation Details}
\label{app:impdetails}
\subsection{Training Configuration}

We build SG-PVR on Qwen3.5-9B~\cite{qwen3.5} and train it in two stages.
Stage 1 fine-tunes the base model for scene graph generation, and Stage 2 continues from the resulting Stage 1 checkpoint to learn the full reward reasoning behavior.
Both stages use LoRA~\cite{hu2022lora} with rank 64 and alpha 128, applied to the linear layers of the language model.
Videos are uniformly sampled at 16 frames per clip in both stages.
We use the AdamW optimizer with a constant learning rate schedule after a 3\% warmup, weight decay of 0.01, and gradient clipping of 1.0. 
We do not perform a systematic hyperparameter search; the reported hyperparameters are fixed across experiments.
All models are trained in bfloat16 precision.
For the three-run statistical robustness analysis, we use random seeds \{42, 43, 44\}.

\textit{Stage 1} is trained for 2 epochs on 6~NVIDIA H100 (80GB) GPUs with a per-device batch size of 1, 2 gradient accumulation steps, a learning rate of $2 \times 10^{-5}$, and a maximum sequence length of 4096. 
Training takes approximately 12 hours in total.

\textit{Stage 2} is trained for 5 epochs on 8~NVIDIA H100 (80GB) GPUs, with a per-device batch size of 1, 8 gradient accumulation steps, a learning rate of $5 \times 10^{-5}$, and a maximum sequence length of 6144.
Training takes approximately 8 hours in total.

% Data construction
\subsection{Stage 1: Scene Graph Supervision}
\label{sec:stage1_data}

\paragraph{Source Dataset.} 
We use Synthetic Visual Genome 2 (SVG2)~\cite{gao2026synthetic} as the base corpus for scene graph supervision. 
SVG2 is a large-scale panoptic video scene graph dataset containing over 636K videos with 6.6M object instances, 52.0M attributes, and 6.7M temporally grounded relations, sourced from the SA-V (43K) and PVD (593K) video collections. 
Annotations are produced by an automated pipeline combining multi-scale panoptic segmentation, trajectory tracking, and GPT-5-based spatio-temporal relation inference; human verification reports 93.8\%, 88.3\%, and 85.4\% accuracy on objects, attributes, and relations respectively, indicating sufficient annotation reliability for use as supervision.

\paragraph{Annotation Refinement Procedure.}
Although SVG2 annotations are reliable at the entity-attribute-relation level, they are exhaustive and include incidental background entities and ambient relations that are not semantically central to the video. 
Such density is appropriate for scene-graph generation as a standalone task, but excessive for our purpose, where the scene graph serves as a lookup target for prompt-level claim verification and benefits from focusing on the video's primary semantic content. 
We therefore use Qwen3.5-27B to filter each scene graph conditioned on the video and its accompanying prompt.
The model retains only annotated entities and relations that are visually supported and semantically central, while discarding incidental items and annotations clearly contradicted by the video.
It does not create or modify annotated entities or relations, and all attributes associated with retained entities are preserved.
The prompt is used only for offline annotation refinement; Stage 1 itself is trained on $(V,\mathcal{G})$ pairs.
For videos in the SA-V split (approximately 6.8\% of SVG2), which are not paired with a textual description, Qwen3.5-27B first generates a descriptive caption, after which the same selection procedure is applied.
This filtering reduces the average number of entities per video from 10.2 to 4.1 and the average number of relations from 11.9 to 3.9, concentrating the supervision signal on the semantic core of each video.
Our model takes a fixed number of 16 input frames per video. 
To ensure each video is sampled at no less than 1 fps, we retain only videos of 16 seconds or shorter.
The resulting Stage 1 training set comprises approximately 307K (video, scene graph) pairs.

\begin{table}[t!]
  \centering
  \small
  \begin{tabular}{lrr}
    \toprule
    \textbf{Dataset} & \textbf{\# Samples} & \textbf{Ratio (\%)} \\
    \midrule
    Q-Eval-100K & 22,067 & 26.5\\
    VideoFeedback & 18,420& 22.1\\
    VideoFeedback2 & 17,614& 21.1\\
    VisionRewardDB-Video& 11,526& 13.8\\
    HVEval & 10,678& 12.8\\
    LiFT-HRA &  2,464 &  3.0\\
    Text2Video-Human &  649 &  0.8\\
    \midrule
    \textbf{Total}                                         & 83,418& \textbf{100.0} \\
    \bottomrule
  \end{tabular}
  \caption{Distribution of Stage 2 training samples across the seven source datasets.}
  \label{tab:dataset_sources}
\end{table}

\subsection{Stage 2: Reasoning Supervision}
\label{sec:stage2_data}

\subsubsection{Training Data}

\paragraph{Source Dataset Composition.} 
Our training corpus comprises 83,418 samples aggregated from seven publicly available video generation preference datasets, namely VideoFeedback~\cite{he2024videoscore}, VideoFeedback2~\cite{he2025videoscore2}, VisionRewardDB-Video~\cite{xu2026visionreward}, Text2Video-Human~\cite{rapidata_t2v_human_preferences_2025}, LiFT-HRA~\cite{wang2024lift}, Q-Eval-100K~\cite{zhang2025q}, and HVEval~\cite{wu2025hveval}. 
As summarized in Table~\ref{tab:dataset_sources}, the corpus is dominated by four large-scale sources---Q-Eval-100K (26.5\%), VideoFeedback (22.1\%), VideoFeedback2 (21.1\%), and VisionRewardDB-Video (13.8\%)---collectively accounting for 83.5\% of training samples. 
The remaining 16.5\% is contributed by HVEval, LiFT-HRA, and Text2Video-Human.
Each source dataset was selected on the basis of public availability and the provision of pointwise dimension-level scores.

% \begin{figure}[!t]
%   \centering
%   \includegraphics[width=\columnwidth]{figures/dataset_distribution.pdf}
%   \caption{Distribution of training samples across the seven source datasets.}
%   \label{fig:dataset_dist}
% \end{figure}

\paragraph{Score Normalization.}
The seven source datasets employ heterogeneous annotation protocols with varying score ranges and granularities.
Where a dataset provides a score for a given dimension (\ding{51}\ in Table~\ref{tab:score_norm}), we use the source-provided label after mapping it to a common 1--5 integer scale.
Ordinal scores are converted using fixed mappings, while continuous scores are either rounded or discretized according to the source dataset.
VisionRewardDB-Video provides multiple fine-grained sub-dimensions rather than one score per target dimension.
We use \textit{alignment} for SA; aggregate \textit{clear} and \textit{detail refinement} for VQ; aggregate \textit{image quality stability}, \textit{shape throughout}, and \textit{movement smoothness} for TQ; and aggregate \textit{physics law} and \textit{movement reality} for PC.
The selected sub-dimension scores are first rescaled from their native ranges to 1--5, averaged within each target dimension, and rounded to the nearest integer.
For dimensions not covered by a dataset's original annotations (\ding{55}), we generate pseudo-scores using VQ-Insight~\cite{zhang2026vq} for VQ and TQ, and VideoPhy-2-AutoEval~\cite{bansal2025videophy} for PC.
VQ-Insight produces continuous scores on a 0--100 scale, which we linearly rescale to 1--5.
VideoPhy-2-AutoEval directly outputs integer scores on the 1--5 scale, requiring no further conversion.
Both models were selected for their strong pointwise scoring performance on the corresponding dimensions at the time of data construction.
Table~\ref{tab:score_norm} summarizes the source-specific normalization rules.

\begin{table*}[t!]
  \centering
  \small
  \setlength{\tabcolsep}{4pt}
  \begin{tabular*}{\textwidth}{@{\extracolsep{\fill}}llccccl@{}}
    \toprule
    \textbf{Dataset} & \textbf{Original Scale}
      & \textbf{SA} & \textbf{VQ} & \textbf{TQ} & \textbf{PC}
      & \textbf{Normalization} \\
    \midrule
    VideoFeedback
      & 1--4 integer
      & \ding{51} & \ding{51} & \ding{51} & \ding{51}
      & $1{\to}1,\ 2{\to}2,\ 3{\to}4,\ 4{\to}5$ \\
 
    VideoFeedback2
      & 1--5 integer
      & \ding{51} & \ding{51} & \ding{55} & \ding{51}
      & as-is \\
 
    VisionRewardDB-Video
      & varied discrete scales
      & \ding{51} & \ding{51} & \ding{51} & \ding{51}
      & selected sub-dimensions $\to$ rescale $\to$ mean $\to$ round \\

    Text2Video-Human
      & 0--1 continuous
      & \ding{51} & \ding{55} & \ding{51} & \ding{55}
      & 5 equal-width bins \\
 
    LiFT-HRA
      & 3-level
      & \ding{51} & \ding{51} & \ding{51} & \ding{55}
      & $\text{Bad}{\to}1,\ \text{Normal}{\to}3,\ \text{Good}{\to}5$ \\
 
    Q-Eval-100K
      & 1--5 continuous
      & \ding{51} & \ding{51} & \ding{55} & \ding{55}
      & round to integer \\
 
    HVEval
      & 0--100 continuous
      & \ding{51} & \ding{51} & \ding{51} & \ding{55}
      & 5-bin uniform \\
    \bottomrule
  \end{tabular*}
  \caption{%
    Score normalization applied to each source dataset.
    \ding{51}~denotes a source-provided label;
    \ding{55}~denotes a dimension absent from the source dataset and filled with a pseudo-score.
  }
  \label{tab:score_norm}
\end{table*}

\begin{figure}[t!]
  \centering
  \includegraphics[width=\columnwidth]{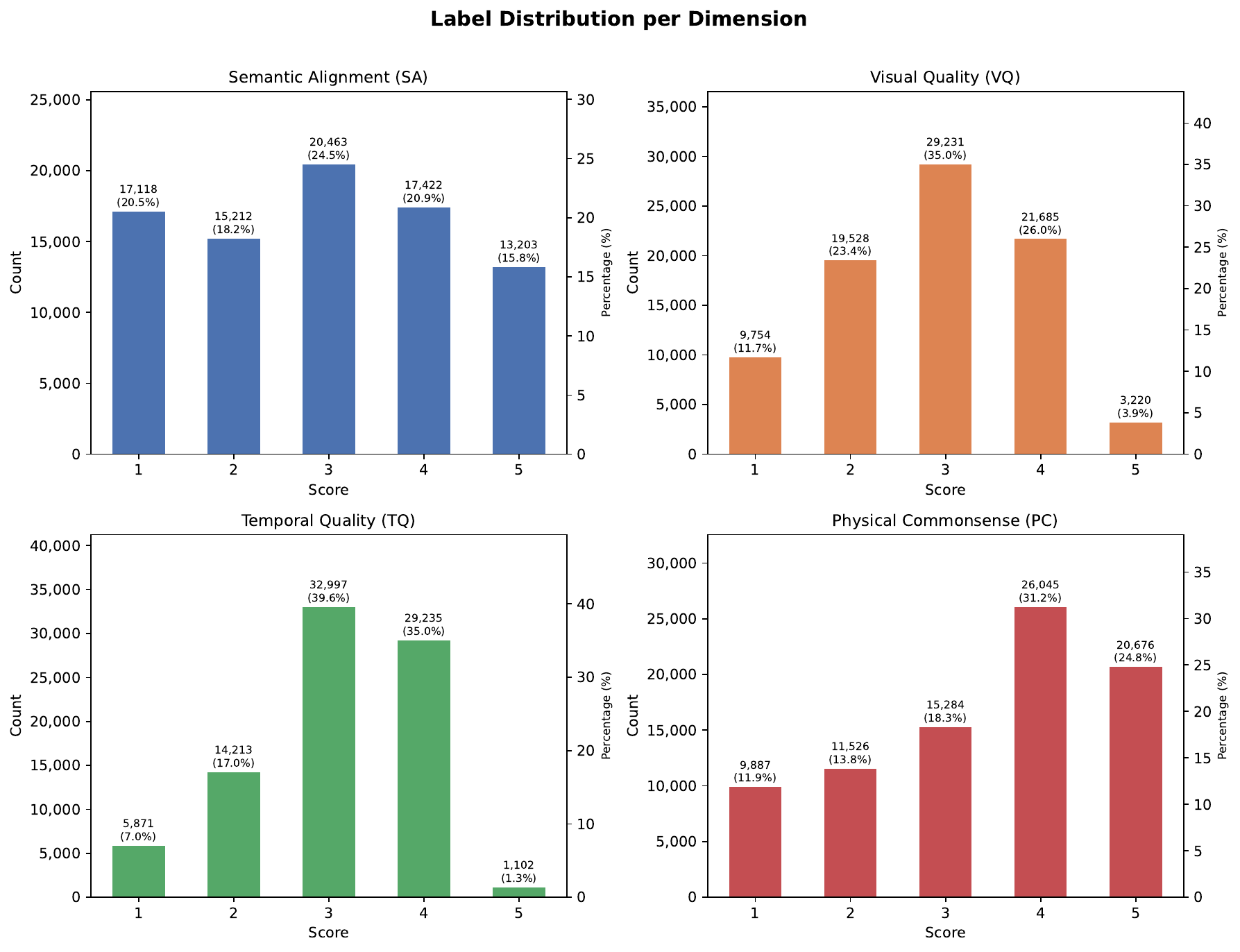}
  \caption{%
    Score label distributions for Semantic Alignment (SA), Visual Quality (VQ), Temporal Quality (TQ), and Physical Commonsense (PC).
  }
  \label{fig:label_dist}
\end{figure}

\paragraph{Final Training Set and Label Distribution.}
For each source video, we combine the Stage 1-generated scene graph with semantic alignment and quality reasoning traces constructed using the procedures below.
We retain a sample only when both reasoning constructions succeed and combine all three components into the Stage 2 supervision target.
This results in 83,418 final training samples.
Figure~\ref{fig:label_dist} shows the resulting score distributions for Semantic Alignment (SA), Visual Quality (VQ), Temporal Quality (TQ), and Physical Commonsense (PC).
The distributions vary across dimensions due to the heterogeneous annotation protocols and label sources described above.

\subsubsection{Semantic Reasoning Trace Construction}
\label{sec:sa_trace_construction}

\begin{algorithm*}[!t]
\small
\vspace{0.6em}
\caption{Semantic Alignment Reasoning Trace Construction}
\label{alg:sa_trace_construction}
\begin{algorithmic}[1]
\Statex \textbf{Notation:} Let $\mathcal{D}$ be the source training set, where each sample is denoted as $(P, V, \mathcal{G}, s^{gt})$.
Here, $P$ is the prompt, $V$ is the video, $\mathcal{G}$ is the spatio-temporal scene graph, and $s^{gt}$ is the ground-truth semantic alignment score.
The verification plan is $\mathcal{P}=\{(c_i,m_i)\}_{i=1}^{N}$, where $c_i$ is a claim and $m_i \in \{\textsc{Critical}, \textsc{Minor}\}$ is its importance label.
A reasoning trace $r$ consists of claim-level evidence and judgments
$j_i \in \{\textsc{Supported}, \textsc{PartiallySupported}, \textsc{Contradicted}\}$,
a final analysis, and a predicted score $s(r)$.
We generate $K=5$ candidates per sample and allow at most $T_f=2$ format refinement rounds.

\State Initialize the final semantic reasoning dataset
$\mathcal{D}_{\mathrm{SA}} \gets \emptyset$.

\For{each source sample $(P,V,\mathcal{G},s^{gt}) \in \mathcal{D}$}

    \State \textbf{Step 1. Verification plan generation:}
    Generate $\mathcal{P}$ from $P$, where $\mathcal{P}$ contains
    claim--importance pairs $(c_i,m_i)$.

    \State \textbf{Step 2. Generate candidate SA reasoning traces:}
    Generate a set $\mathcal{R}$ of $K$ candidate reasoning traces from
    $P$, $V$, $\mathcal{G}$, and $\mathcal{P}$.

    \State Initialize the accepted trace set
    $\mathcal{R}_{\mathrm{acc}} \gets \emptyset$.

    \State \textbf{Step 3. Format filtering and refinement:}

    \For{each candidate trace $r \in \mathcal{R}$}

        \State $t \gets 0$
        \State $f \gets \text{\textsc{FormatCheck}}(r,\mathcal{P})$

        \While{$f=\text{\textsc{Fail}}$ \textbf{and} $t<T_f$}
            \State Refine $r$ only to fix format violations.
            \State $t \gets t+1$
            \State $f \gets \text{\textsc{FormatCheck}}(r,\mathcal{P})$
        \EndWhile

        \If{$f=\text{\textsc{Fail}}$}
            \State Discard $r$.
            \State \textbf{continue}
        \EndIf

        \State \textbf{Step 4. Score matching and consistency verification:}

        \If{$s(r)=s^{gt}$}
            \State Apply Consistency Verification to $r$.

            \If{$r$ passes Consistency Verification}
                \State Add $r$ to $\mathcal{R}_{\mathrm{acc}}$.
                \State \textbf{continue}
            \Else
                \State \textbf{Consistency-guided refinement:}
                Refine $r$ into $r'$ using consistency verification feedback,
                $s^{gt}$, $P$, $V$, $\mathcal{G}$, and $\mathcal{P}$.
            \EndIf

        \Else
            \State \textbf{GT-score-guided regeneration:}
            Regenerate a new trace $r'$ using
            $s^{gt}$, $P$, $V$, $\mathcal{G}$, and $\mathcal{P}$.
        \EndIf

        \State \textbf{Step 5. Final filtering:}
        Apply format filtering, exact GT-score matching, and
        Consistency Verification to $r'$.

        \If{$r'$ passes all three filters}
            \State Add $r'$ to $\mathcal{R}_{\mathrm{acc}}$.
        \EndIf

    \EndFor

    \State \textbf{Step 6. Final trace selection:}
    \If{$\mathcal{R}_{\mathrm{acc}} \neq \emptyset$}
        \State Randomly sample one trace
        $\tilde{r} \sim \mathrm{Uniform}(\mathcal{R}_{\mathrm{acc}})$.
        \State Add $\tilde{r}$ to $\mathcal{D}_{\mathrm{SA}}$.
    \EndIf

\EndFor

\State \Return $\mathcal{D}_{\mathrm{SA}}$

\end{algorithmic}
\end{algorithm*}

Algorithm~\ref{alg:sa_trace_construction} summarizes the construction pipeline for semantic alignment reasoning traces.
For each source sample, we generate $K=5$ candidate traces and process each candidate independently through format checking, score matching, and consistency verification.
Candidates that satisfy the required checks, either directly or after regeneration/refinement, are collected into an accepted set.
If multiple traces are accepted for the same source sample, we randomly retain one as the final semantic reasoning supervision.
Below, we describe the main components of this procedure.

\begin{table*}[t!]
\centering
\small
\begin{tabular}{lcccc}
\toprule
\textbf{Trace condition}
& \shortstack{\textbf{Evidence}\\\textbf{Grounding (\%)}}
& \shortstack{\textbf{Claim-Scope}\\\textbf{Alignment (\%)}}
& \shortstack{\textbf{Judgment}\\\textbf{Validity (\%)}}
& \shortstack{\textbf{Judgment}\\\textbf{Consistency (\%)}} \\
\midrule
Before GT conditioning
& 82.34 & 90.75 & 84.63 & 84.57 \\
After GT conditioning
& 79.87 & 88.46 & 82.59 & 85.71 \\
\midrule
Regression rate
& 13.24 & 8.83 & 11.04 & 5.71 \\
\bottomrule
\end{tabular}
\caption{
Reasoning-trace faithfulness before and after GT-conditioned refinement/regeneration on the same 350 paired examples.
Faithfulness regression rate denotes the proportion of reasoning components that satisfy a given criterion before GT conditioning but fail to satisfy the same criterion afterward.
}
\label{tab:gt-anchoring-analysis}
\end{table*}

\paragraph{Format Filtering and Refinement.}
We first apply automatic format filtering to each candidate reasoning trace to ensure that it preserves the structure of its verification plan.
A candidate trace is considered format-valid only if it satisfies three requirements:
(i) \emph{plan-structure alignment}, where the trace follows the predefined semantic criteria and matches the criterion and claim structure of the verification plan;
(ii) \emph{claim-level verification validity}, where each claim has one evidence-grounded evaluation with a valid judgment label; and
(iii) \emph{final-output compliance}, where the final analysis is placed after all claim-level evaluations and the final score follows the required format.
If a candidate violates any requirement, we provide error-specific feedback and perform format refinement for up to $T_f=2$ rounds.
During refinement, the model receives the original prompt, verification plan, video $V$, scene graph $\mathcal{G}$, the pre-refinement trace, and feedback describing the detected structural violations.
The pre-refinement trace is used only to identify the failure mode.
To avoid superficial local edits, the model regenerates the full trace while preserving the claim order, criterion headings, and one-to-one correspondence between claims and evaluations.
Candidates that remain format-invalid after the allowed refinement rounds are discarded.

\paragraph{Score Matching and Consistency Verification.}
For each format-valid candidate, we first compare its predicted semantic alignment score with the ground-truth score.
If the scores match, we apply consistency verification to determine whether the accepted score is supported by a coherent reasoning process.
We verify two aspects:
\emph{judgment consistency}, which checks whether claim-level judgments are logically compatible with each other, and
\emph{rubric-score consistency}, which checks whether the final score follows the predefined scoring rubric given the distribution of Supported, Partially Supported, and Contradicted judgments over Critical and Minor claims.
For example, if an entity is judged as Contradicted, claims about that entity's action, attribute, spatial relation, or temporal behavior should generally not be judged as Supported.
A score-matched trace that passes consistency verification is directly added to the accepted trace set.

\paragraph{Score-Guided Regeneration and Consistency-Guided Refinement.}
Candidates that do not immediately pass the preceding checks are corrected according to their failure type.
For traces whose predicted scores do not match the ground-truth score, we perform GT-score-guided regeneration by providing the ground-truth score together with the original prompt, verification plan, video $V$, and scene graph $\mathcal{G}$.
For traces whose scores match the ground truth but fail consistency verification, we instead perform consistency-guided refinement by providing the original trace, consistency feedback, and the same conditioning information.
In both cases, the ground-truth score serves as the target score that must be justified by claim-level evidence and the scoring rubric.
The corrected trace is then subjected to final filtering, which re-applies format validation, exact GT-score matching, and consistency verification.
Only corrected traces that pass all three checks are added to the accepted trace set.

\paragraph{Final Trace Selection.}
After all $K$ candidates for a source sample have been processed, we randomly select one trace from the accepted set, if non-empty, and retain it as the semantic reasoning supervision for that source sample.
Thus, each retained source sample contributes exactly one semantic alignment reasoning trace to the final training set.

\paragraph{Effect of GT-Anchored Refinement and Regeneration.}
As described above, traces with incorrect predicted scores are regenerated using the ground-truth (GT) score as an anchor, while traces that fail consistency verification are refined using both consistency feedback and the GT score.
We therefore examine whether GT-anchored refinement and regeneration preserve faithful, evidence-grounded reasoning rather than rationalizing the target score with unsupported information.

We compare paired traces from the same 350 examples before and after GT conditioning.
The macro-averaged claim-level judgment flip rate is 18.49\%, showing that conditioning revises a subset of verification judgments rather than broadly changing the claim-level decisions.
We next examine whether these changes compromise reasoning-trace faithfulness.

Table~\ref{tab:gt-anchoring-analysis} evaluates the paired traces using the four faithfulness criteria from Appendix~\ref{sec:reasoning-trace}.
All aggregate metrics change by less than 2.5 percentage points after GT conditioning, indicating that overall reasoning-trace faithfulness remains largely stable.
To capture localized degradations that may be obscured by aggregate scores, we additionally report faithfulness regression rates, computed at the claim level for Evidence Grounding, Claim-Scope Alignment, and Judgment Validity, and at the trace level for Judgment Consistency.
The regression rates range from 5.71\% to 13.24\%, meaning that 86.76--94.29\% of initially criterion-satisfying reasoning components continue to satisfy the corresponding criterion after GT conditioning.

Overall, these results indicate that GT-anchored refinement and regeneration can revise individual verification judgments while largely preserving evidence grounding and reasoning-trace faithfulness, supporting their use for constructing score-corrected and consistency-refined reasoning supervision.

\subsubsection{Quality Reasoning Trace Construction}
For each video, we use Qwen3.5-27B to sample $N=6$ candidate traces, each jointly producing rationales and 1--5 scores for Visual Quality (VQ), Temporal Quality (TQ), and Physical/Common-sense Consistency (PC).
For each dimension, we use the source-provided GT score when available and otherwise the corresponding pseudo-score as the target score.

\paragraph{Rationale Selection and Validation.}
A candidate rationale is considered score-matched for a dimension if its predicted score is within $\pm1$ of the target score. 
Because a single candidate trace does not always provide score-matched rationales for all three dimensions, we select rationales at the dimension level across the sampled candidates. 
Whenever possible, all three rationales are taken from the same trace; otherwise, we prioritize a trace that matches the largest number of dimensions and fill only the remaining dimensions from other candidates.

We additionally check whether each selected rationale is consistent with the score it originally predicts. 
Specifically, we mask its numerical score and independently re-infer the score from the rationale alone. 
We retain the rationale only when the re-inferred score matches its original predicted score. 
This check prevents a rationale from being accepted solely because its numerical prediction is close to the target while its textual reasoning supports a different score.
For candidate rationales that pass this consistency check, GPT-4o-mini lightly adjusts the wording to align with the target score, following VideoScore2~\cite{he2025videoscore2}. 
This adjustment only softens or intensifies existing observations and does not introduce new visual claims.

\paragraph{GT-Conditioned Regeneration.}
If no valid candidate rationale remains for a dimension after score matching and consistency validation, we regenerate the rationale conditioned on the video and the target score. 
The regeneration prompt explicitly instructs the model to ground its reasoning in concrete visual observations rather than justify the target score abstractly. 
We then apply the same rationale--score consistency check by masking the generated numerical score and re-inferring it from the rationale alone. 
If the regenerated rationale fails this check, we discard the video; only videos with valid rationales for all three quality dimensions are retained.

\section{Evaluation Details}
\label{app:evaldetails}
\subsection{Benchmarks}
\label{app:benchmarks}

\paragraph{VideoScore-Bench-v2~\cite{he2025videoscore2}.}
VideoScore-Bench-v2 is the 500-video held-out test split of VideoFeedback2, covering videos generated by 22 T2V models spanning early baselines through recent state-of-the-art systems.
Each video carries human-assigned scores (1--5) across three dimensions---Visual Quality, Text Alignment, and Physical/Common-Sense Consistency---with high inter-annotator agreement, accompanied by CoT-style rationales expanded from annotator comments via an LLM-based pipeline.
Prompt coverage includes multi-action, OCR, and camera-motion scenarios designed to probe known failure modes of T2V generators.

\paragraph{LGVQ~\cite{zhang2025benchmarking}.}
LGVQ (Large-scale Generated Video Quality assessment) consists of 2,808 AI-generated videos produced by 6 T2V models from 468 curated prompts, with mean opinion scores (MOS) collected from 60 subjects across three independently evaluated dimensions: Spatial Quality, Temporal Quality, and Text-Video Alignment.
% A systematic benchmark of existing metrics revealed that none adequately assessed all three dimensions simultaneously, making LGVQ a discriminative testbed for multi-dimensional video quality evaluators.

\paragraph{MJ-Bench-Video~\cite{tong2025mj}.}
MJ-Bench-Video is a large-scale video preference benchmark comprising 10,842 videos (5,421 preference pairs) annotated across five aspects---Alignment, Safety, Fineness, Coherence \& Consistency, and Bias \& Fairness---with 28 fine-grained criteria.
Although originally designed for pairwise preference evaluation, each video also carries per-video aspect-level and overall scores on a $\{0,1,2\}$ scale, which allows the benchmark to be repurposed for pointwise scoring, as done in \citet{he2025videoscore2}.
Following this protocol, we use the held-out test split of 2,170 videos and report Overall accuracy against the per-video overall score, with the dimension mapping and rescaling described in Appendix~\ref{app:dim_mapping}.

\paragraph{Vinoground~\cite{zhang2024vinoground}.}
Vinoground is a temporal counterfactual benchmark of 500 counterfactual pairs built from 1,000 natural videos. 
Each pair consists of two videos and two captions, where the negative caption is composed of the exact same words as the positive caption reordered to describe a temporally swapped event sequence, and each caption matches exactly one of the two videos.
Even frontier LMMs perform far below the human baseline on this benchmark.
We use Vinoground as a source for our temporal semantics test set, filtering out instances solvable by static cues alone and sampling 300 of the remaining pairs; the filtering criteria are described in Appendix~\ref{app:temporal_details}.
Our evaluation protocol is the direct analogue of Vinoground's video score: for each caption, the model must identify the matching video from the counterfactual pair, and an instance counts as correct only when both directions are judged correctly.

\paragraph{TimeBlind~\cite{li2026timeblind}.}
TimeBlind is a diagnostic benchmark for compositional spatio-temporal understanding.
Each instance pairs two videos sharing near-identical static content but differing solely in a targeted temporal dimension, with complementary questions whose ground-truth answers flip between videos to neutralize language priors.
Its taxonomy spans atomic event recognition (\textit{Events}), event property characterization (\textit{Event Attributes}), and inter-event dependency reasoning (\textit{Structural Event Logic}), and the best-performing MLLM reaches only 48.2\% Instance Accuracy against a 98.2\% human baseline.
We use all 300 unique video pairs, converting the original question-form prompts into declarative descriptions for alignment scoring, with the conversion procedure described in Appendix~\ref{app:temporal_details}. Our Overall metric mirrors TimeBlind's Instance Accuracy, counting an instance correct only when both videos in the pair are correctly discriminated.

\paragraph{T2V-CompBench~\cite{sun2025t2v}.}
T2V-CompBench is a benchmark for compositional text-to-video generation comprising 1,400 prompts across seven dimensions, including Consistent Attribute Binding, Dynamic Attribute Binding, Spatial Relationships, Motion Binding, Action Binding, Object Interactions, and Generative Numeracy.
It serves as the benchmark for our test-time reranking evaluation.
T2V-CompBench provides dimension-specific evaluation metrics based on multimodal language models, object detection, and tracking, whose agreement with human judgments is validated in the original study.
We use the official per-dimension evaluators to assess the videos selected by each reward model.

\subsection{Baseline Models}
\label{app:baselines}

\paragraph{OmniScore~\cite{liu2025videodpo}.}
OmniScore outputs a single continuous pointwise score that aggregates visual quality, covering both intra-frame fidelity and inter-frame temporal consistency, together with text-video semantic alignment, so that no preference aspect is neglected in a single scalar reward.
It is introduced in VideoDPO, which pioneers Direct Preference Optimization for text-to-video diffusion models: OmniScore automatically constructs preference pairs without human labeling, a re-weighting strategy assigns higher DPO weights to pairs with clearer score distinctions, and the pipeline is validated on 3 open-source T2V models.

\paragraph{VBench-2.0~\cite{zheng2025vbench}.}
VBench-2.0 reports model-level scores over its own prompt suite across 18 fine-grained dimensions grouped under Human Fidelity, Controllability, Creativity, Physics, and Commonsense, rather than scoring individual prompt-video pairs.
Its evaluation combines generalist VLMs and LLMs for semantic dimensions with specialist anomaly detectors where generalists are unreliable, such as human anatomy and motion rationality, and the resulting per-dimension scores are validated against extensive human preference annotations.
To use it as an instance-level baseline, we adapt its VQA-based per-dimension pipeline to score individual prompt-video pairs, as described in Appendix~\ref{app:semantic_dim_mapping}.

\paragraph{VideoScore~v1.1~\cite{he2024videoscore}.}
VideoScore~v1.1 outputs pointwise scores on a 1--4 scale over five dimensions, namely Visual Quality, Temporal Consistency, Dynamic Degree, Text-to-Video Alignment, and Factual Consistency, and pairwise preferences are derived by comparing averaged scores.
It is built on Mantis-Idefics2-8B with the language model head replaced by a regression layer, trained on VideoFeedback, a dataset of 37.6K human-annotated videos from 11 T2V models.

\paragraph{VisionReward~\cite{xu2026visionreward}.}
VisionReward outputs a continuous pointwise reward by decomposing visual preference into a hierarchy of binary yes/no judgment questions and linearly weighting the responses, and also supports pairwise comparison.
It fine-tunes CogVLM2-Video-13B on 33K annotated videos covering 2M binary question-answer pairs.

\paragraph{VideoReward~\cite{liu2025improving}.}
VideoReward outputs per-dimension continuous scores over Visual Quality, Motion Quality, and Text Alignment, and supports pairwise comparison.
It trains a Qwen2-VL-2B backbone on 182K human-labeled pairwise comparisons collected from 12 modern T2V models.

\paragraph{LiFT~\cite{wang2024lift}.}
LiFT-Critic outputs pointwise scores over Semantic Consistency, Motion Smoothness, and Video Fidelity, without support for pairwise comparison.
It is built on VILA-1.5-13B and trained on LiFT-HRA, a dataset of 10K human annotations each pairing a numerical rating with a free-text rationale; the resulting scores serve as reward weights for T2V fine-tuning via reward-weighted likelihood maximization.

\paragraph{UnifiedReward~2.0~\cite{wang2025unified}.}
UnifiedReward supports both pairwise ranking and pointwise scoring, and is the first reward model trained jointly on image and video generation and understanding tasks, demonstrating cross-task synergy between the two.
UnifiedReward~2.0 is a strengthened checkpoint of this framework that extends scoring to explicit dimensions of Alignment, Coherence/Physics, and Style for both pairwise and pointwise evaluation.

\paragraph{VideoScore2~\cite{he2025videoscore2}.}
VideoScore2 outputs pointwise scores as normalized floats in the 1--5 range over Visual Quality, Text Alignment, and Physical Consistency, each accompanied by a chain-of-thought analysis.
It is built on Qwen2.5-VL-7B-Instruct and trained on VideoFeedback2, a dataset of 27.2K videos with human scores and CoT rationales from over 20 T2V models, via SFT cold-start for structured output formatting followed by GRPO-based RL with an accuracy reward penalizing per-dimension deviations exceeding one point.

\paragraph{UnifiedReward-Think~\cite{wang2025unifiedthink}.}
UnifiedReward-Think operates in pairwise comparison mode, generating an explicit long chain-of-thought rationale before emitting its final preference decision.
It extends UnifiedReward through a three-stage pipeline of cold-start distillation from GPT-4o, rejection sampling on correctly reasoned samples, and GRPO-based reinforcement fine-tuning on incorrectly predicted ones.

\paragraph{UnifiedReward-Flex~\cite{wang2026unified}.}
UnifiedReward-Flex operates in pairwise comparison mode as a unified personalized reward model: rather than applying fixed evaluation rubrics, it dynamically instantiates fine-grained criteria by interpreting the semantic intent of the prompt and grounding on visual evidence, making it sensitive to content-specific cues that static models overlook.
It is trained in two stages, SFT on reasoning traces distilled from closed-source VLMs followed by DPO on curated preference pairs to strengthen reasoning fidelity and discriminative alignment.

\subsection{Dimension Mapping for Pointwise Benchmarks}
\label{app:dim_mapping}
 
Since baseline models adopt varying output dimensions and scoring scales, we align their outputs with the ground-truth dimensions of each pointwise benchmark via semantic correspondence, following the dimension descriptions defined in each model's original paper.
For models that do not cover a given dimension (marked ``---''), evaluation on that dimension is excluded.
Although score broadcasting can fill otherwise undefined dimensions, we report only semantically corresponding native outputs to avoid interpreting a single composite reward as multiple distinct dimensions.

\paragraph{VideoScore-Bench-v2.}
VideoScore-Bench-v2 defines three ground-truth dimensions: Visual Quality (V), Text Alignment (A), and Physical/Common-sense Consistency (P).
We report per-dimension Accuracy, the exact-match rate between the rounded prediction and the ground-truth integer score; Relaxed Accuracy, which counts predictions within one point of the ground truth as correct; and PLCC, the Pearson linear correlation computed on raw float predictions.
For all baselines, we follow the dimension mapping and score rescaling of \citet{he2025videoscore2}: in particular, for models without a fixed output range such as VideoReward, raw scores are assumed to approximately follow a Gaussian distribution and are discretized into $\{1,\dots,5\}$ using the thresholds $\Phi^{-1}(0.2)$, $\Phi^{-1}(0.4)$, $\Phi^{-1}(0.6)$, and $\Phi^{-1}(0.8)$ of the standard normal CDF.
When the description in Appendix B.2 of \citet{he2025videoscore2} differs from its official implementation, we follow the released code.
For VideoScore~v1.1, we quote the results reported in \citet{he2025videoscore2}, which does not document its dimension mapping or rescaling.
Table~\ref{tab:mapping_vs2} summarizes the dimension mapping and score handling for the baselines we evaluate.
 
\begin{table*}[t!]
\centering
\resizebox{\textwidth}{!}{%
\begin{tabular}{lll}
\toprule
\textbf{Model} & \textbf{Mapped Dimensions (V, P, A)} & \textbf{Rescaling Method} \\
\midrule
VideoScore~v1.1 & visual quality, factual consistency, text-video alignment& No rescaling \\
VideoReward & visual quality, ---, text alignment & Gaussian quintile $\to[1,5]$ \\
LiFT & video fidelity, ---, semantic consistency & Map \{Bad, Normal, Good\} $\to\{1,3,5\}$ \\
UnifiedReward~2.0 & --- , physical, alignment & No rescaling \\
VideoScore2 & visual quality, physical consistency, text alignment & No rescaling \\

\bottomrule
\end{tabular}
}
\caption{Dimension mapping and score handling for VideoScore-Bench-v2.
         ``---'': no corresponding output; excluded from evaluation.}
\label{tab:mapping_vs2}
\end{table*}

\begin{table*}[t!]
\centering
\small
\begin{tabular}{llll}
\toprule
\textbf{Model} & \textbf{Spatial (S)} & \textbf{Temporal (T)} & \textbf{Alignment (A)} \\
\midrule
VideoScore~v1.1 & visual quality & temporal consistency & text-video alignment \\
VideoReward & visual quality & --- & text alignment \\
LiFT & video fidelity & motion smoothness & semantic consistency \\
UnifiedReward~2.0 & --- & --- & alignment \\
VideoScore2 & visual quality & --- & text alignment \\
\bottomrule
\end{tabular}
\caption{Dimension mapping for LGVQ.
         ``---'': no corresponding output; excluded from evaluation.}
\label{tab:mapping_lgvq}
\end{table*}
 
\paragraph{LGVQ.}
LGVQ defines three ground-truth dimensions: Spatial Quality (S), Temporal Quality (T), and Text-Video Alignment (A).
We report SRCC and PLCC between model predictions and the mean opinion scores for each dimension; since both are computed on raw continuous predictions, no score rescaling is required.
Note that LGVQ's Temporal Quality dimension targets low-level temporal fidelity artifacts such as motion blur and frame jitter, rather than semantic event structure; we therefore construct a dedicated temporal semantics test set, described in Appendix~\ref{app:temporal_details}, to probe the latter, which existing pointwise benchmarks do not cover.
Table~\ref{tab:mapping_lgvq} summarizes the dimension mapping applied to each baseline.

\paragraph{MJ-Bench-Video.}
MJ-Bench-Video provides aspect-level and overall ground-truth scores on a $\{0,1,2\}$ scale.
Following the protocol of \citet{he2025videoscore2}, we report Overall accuracy: we first average a model’s dimension scores, cast the mean to an integer, and then rescale it to $\{0,1,2\}$ via $1\!\to\!0$, $\{2,3\}\!\to\!1$, $\{4,5\}\!\to\!2$.
The resulting score is compared against the benchmark’s overall score.
For the per-prompt analyses in Figure~\ref{fig:mjbench_4model_compare} and Table~\ref{tab:mjbench_prompt_type}, we use only the \textit{Alignment} aspect, comparing each model's SA score rescaled by the same rule against the aspect-level ground truth.
We evaluate all baselines ourselves under this protocol, applying the dimension mapping and score rescaling of \citet{he2025videoscore2} to each model's outputs.

\subsection{Temporal Semantics Test Set}
\label{app:temporal_details}

\paragraph{Data sources and filtering.}
From TimeBlind~\cite{li2026timeblind}, we use all 300 positive/negative video pairs.
We convert the original question-form prompts into declarative descriptions using GPT-5.4 and manually inspect every converted description to verify that object references and complementary semantics are preserved.
This conversion enables direct evaluation with caption-based alignment scores while retaining the contrastive structure of the original benchmark.
From Vinoground~\cite{zhang2024vinoground}, three annotators inspect all 500 counterfactual pairs and exclude a pair when the main subjects correspond to different individuals or object instances, or when the two videos contain clearly different scenes or backgrounds. 
These filters reduce static shortcuts that could distinguish the videos without temporal reasoning.
A total of 421 pairs satisfy both criteria, from which we sample 300 to match the scale of TimeBlind. 
Table~\ref{tab:temporal_stats} reports the resulting category and duration statistics.

\paragraph{Output-format neutrality.}
Dataset construction does not use model predictions or depend on a particular reward-model output format. 
Scoring-based, pairwise, and VQA-based models are evaluated with separate protocols matched to their native outputs. 
The filtering procedure is therefore independent of whether a model performs pointwise scoring or pairwise comparison.

\paragraph{Release.}
We will release the evaluation code, construction scripts, selected source identifiers, and filtering annotations, enabling the exact source video pairs to be recovered from the official Vinoground and TimeBlind releases.
Original videos and derived captions will not be redistributed.

\paragraph{Dataset statistics.}

% \begin{table*}[t!]
% \centering
% \resizebox{\textwidth}{!}{%
% \begin{tabular}{lrrrrl}
% \toprule
% \textbf{Source} & \textbf{\# Pairs} & \textbf{Min (s)} & \textbf{Max (s)} & \textbf{Mean (s)} & \textbf{Temporal Categories} \\
% \midrule
% Vinoground & 300 & 1.0 & 55.0 & 8.0 & \begin{tabular}[t]{@{}l@{}}
% \textit{Major}: object, action, viewpoint \\
% \textit{Minor}: interaction, cyclical, spatial, contextual
% \end{tabular}\\[0.5em]
% TimeBlind  & 300 & 0.4 & 50.2 & 8.6 & \begin{tabular}[t]{@{}l@{}}
% \textit{Event}: fine-grained action, state transitions \\
% \textit{Event Attribute}: speed, force, magnitude, duration, direction, repetition \\
% \textit{Structural Logic}: temporal topology, causal contingency, cross-event comparison
% \end{tabular} \\
% \bottomrule
% \end{tabular}
% }
% \caption{Statistics of the temporal semantics test set. Vinoground organizes instances into three major categories (object, action, viewpoint) and four minor categories (interaction, cyclical, spatial, contextual) that capture distinct aspects of temporal understanding. TimeBlind adopts a hierarchical three-level taxonomy spanning atomic events, parametric attributes, and structural logic.}
% \label{tab:temporal_stats}
% \end{table*}

\begin{table*}[t!]
\centering
\resizebox{0.7\textwidth}{!}{%
\begin{tabular}{llcccc}
\toprule
\textbf{Source} & \textbf{Category} & \textbf{\# Pairs} & \textbf{Min (s)} & \textbf{Max (s)} & \textbf{Mean (s)} \\
\midrule
\multirow{4}{*}{Vinoground}
 & action       & 157 & 2.0 & 55.0 & 8.2 \\
 & object       &  83 & 1.0 & 30.0 & 7.8 \\
 & viewpoint    &  60 & 2.0 & 15.1 & 7.6 \\
  \cmidrule(l){2-6}
 & \textit{all} & 300 & 1.0 & 55.0 & 8.0 \\
\midrule
\multirow{4}{*}{TimeBlind}
 & Event                  & 108 & 0.5 & 50.2 &  7.8 \\
 & Event Attribute        &  95 & 0.4 & 23.7 &  8.0 \\
 & Structural Event Logic &  97 & 1.7 & 26.2 & 10.2 \\
  \cmidrule(l){2-6}
 & \textit{all}           & 300 & 0.4 & 50.2 &  8.6 \\
\bottomrule
\end{tabular}
}
\caption{Statistics of the temporal semantics test set. Video durations are measured over both videos of each pair. 
Vinoground pairs are grouped by their major category, with cross-cutting minor categories (interaction, cyclical, spatial, contextual) capturing additional aspects of temporal understanding. 
TimeBlind adopts a hierarchical three-level taxonomy spanning atomic events, parametric attributes, and structural event logic.}
\label{tab:temporal_stats}
\end{table*}

The two sources provide complementary coverage of temporal phenomena.
Vinoground focuses on natural counterfactuals involving actions, object states, and viewpoint changes. 
TimeBlind covers atomic events, event attributes, and structural event logic. 
Table~\ref{tab:temporal_stats} reports the number of pairs and video-duration statistics for each category.
In both cases, each instance takes the form of a minimally contrastive pair, as in:
\begin{quote}
\small
\textit{pos: ``a toddler plays around the grass field before he picks up a water bottle
and drinks''} \\
\textit{neg: ``a toddler picks up a water bottle and drinks before he plays around the
grass field''}
\end{quote}

\paragraph{Evaluation protocol.}
For each instance $(v^{+}, v^{-}, p^{+}, p^{-})$, where $v^{+}$ ($v^{-}$) is the temporally positive (negative) video and $p^{+}$ ($p^{-}$) is the corresponding description, we evaluate reward models under three protocol variants depending on their output format.
 
\textbf{Scoring-based models.} For models that produce scalar alignment scores $r(\cdot, \cdot)$, 
we compute the following binary correctness indicators for each instance:
\begin{align}
\text{Task1} &= \mathbf{1}[r(v^+, p^+) > r(v^-, p^+)] \label{eq:task1}\\
\text{Task2} &= \mathbf{1}[r(v^-, p^-) > r(v^+, p^-)] \label{eq:task2}\\
\text{Overall}   &= \mathbf{1}\!\left[
\text{Task1}=1 \wedge \text{Task2}=1
\right] \label{eq:overall}
\end{align}
where Task1 indicates whether a model ranks the positive video higher when given the positive description, and Task2 indicates whether it ranks the negative video higher for the negative description.

\textbf{Pairwise-comparison models.} For models that output a binary preference $\mathrm{pref}(a,b \mid p) \in \{a,b\}$ given two videos and a prompt, we present the videos in the fixed order $(v^{+}, v^{-})$ under both $p^{+}$ and $p^{-}$, and define the following binary correctness indicators
for each instance:
\begin{align}
  \text{Task1} &= \mathbf{1}\!\left[
                      \mathrm{pref}(v^{+}, v^{-} \mid p^{+}) = v^{+}
                      \right] \label{eq:pair_task1}\\
  \text{Task2} &= \mathbf{1}\!\left[
                      \mathrm{pref}(v^{+}, v^{-} \mid p^{-}) = v^{-}
                      \right] \label{eq:pair_task2}\\
  \text{Overall}   &= \mathbf{1}\!\left[
                      \text{Task1}=1 \wedge \text{Task2}=1
                      \right]                    \label{eq:pair_overall}
\end{align}
where Task1 indicates whether the model selects $v^{+}$ given the positive description, and Task2 indicates whether it selects $v^{-}$ given the negative description.

% To control for position bias---a tendency observed in pairwise models to favour whichever video appears first---the presentation order of $(v^{+}, v^{-})$ is randomised with 50\% probability per instance; the reported accuracy is computed over this randomised ordering.

\textbf{VQA-based models.} For models that answer natural-language questions with binary yes/no outputs---such as VBench-2.0, whose per-dimension evaluation relies on VQA---we map each prompt--video combination to a yes/no question and collect four binary indicators for each instance:
\begin{align}
  a_1 &= \mathbf{1}\!\left[
          \mathrm{answer}(q^{+},\,v^{+}) = \texttt{Yes}
          \right] \label{eq:vqa_a1}\\
  a_2 &= \mathbf{1}\!\left[
          \mathrm{answer}(q^{+},\,v^{-}) = \texttt{No}
          \right] \label{eq:vqa_a2}\\
  a_3 &= \mathbf{1}\!\left[
          \mathrm{answer}(q^{-},\,v^{-}) = \texttt{Yes}
          \right] \label{eq:vqa_a3}\\
  a_4 &= \mathbf{1}\!\left[
          \mathrm{answer}(q^{-},\,v^{+}) = \texttt{No}
          \right] \label{eq:vqa_a4}
\end{align}
where $q^{+}$ ($q^{-}$) denotes the yes/no question corresponding to $p^{+}$ ($p^{-}$). 
These yield:
\begin{align}
\text{Task1} &= a_1 a_2 \\
\text{Task2} &= a_3 a_4 \\
\text{Overall} &= \mathbf{1}\!\left[
\text{Task1}=1 \wedge \text{Task2}=1
\right]
\end{align}

\paragraph{Aggregation.}
For each protocol, Task1 accuracy, Task2 accuracy, and Overall accuracy are computed by averaging the corresponding binary correctness indicators over all 600 test instances.
% This minimal-pairs design ensures that a model can score above chance only by genuinely discriminating temporal structure, not by exploiting object-level or linguistic shortcuts.
This minimal-pairs design is intended to reduce object-level and linguistic shortcuts, making successful discrimination depend primarily on the targeted temporal structure.

\subsection{Dimension Mapping for Temporal Semantics Evaluation}
\label{app:semantic_dim_mapping}

Scoring-based reward models are evaluated on the temporal semantics test set using a single score per prompt-video pair.
For models with multi-dimensional outputs, we use the dimension corresponding to text-video semantic alignment, and the remaining quality dimensions are not used.
VBench-2.0~\cite{zheng2025vbench} is instead evaluated under the VQA-based protocol of Appendix~\ref{app:temporal_details}. 
The rest of this section describes how test instances are mapped into VBench-2.0's question templates.

We map instances from Vinoground~\cite{zhang2024vinoground} and TimeBlind~\cite{li2026timeblind} onto three of its temporal controllability dimensions where applicable, namely Dynamic Attribute, Dynamic Spatial Relationship, and Motion Order Understanding.
All VQA inference uses Qwen3.5 as the fixed backbone, following VBench-2.0's video-based multi-question answering pipeline.

\paragraph{Vinoground.}
Vinoground's category labels are coarse and do not align directly with VBench-2.0's dimension taxonomy, so we perform prompt-level dimension assignment via GPT that identifies the \textit{primary temporal change signal} of each caption and assigns one of the three target dimensions, following a fixed decision hierarchy with explicit tie-breaking rules.
Once a dimension is assigned, auxiliary yes/no questions are generated from the positive caption using VBench-2.0's question templates.

A subset of Vinoground prompts is camera-centric, where the primary temporal change concerns the camera's focus or viewpoint rather than a scene entity. 
As VBench-2.0's Camera Motion dimension targets physical camera movements (pan, zoom, tilt) rather than the temporal ordering of camera attention, we evaluate these prompts using the same three-question yes/no template as Dynamic Attribute and Dynamic Spatial Relationship, with the camera treated as the entity of interest.

\paragraph{TimeBlind.}
TimeBlind's fine-grained sub-dimension labels map directly onto VBench-2.0 dimensions, with \textbf{State Transition} mapped to Dynamic Attribute and \textbf{Direction} mapped to Dynamic Spatial Relationship.
For instances whose temporal reasoning involves higher-level causal or structural dependencies that do not correspond to any of the three VBench-2.0 dimensions, we retain TimeBlind's original binary yes/no questions directly for VQA.

\begin{table*}[t!]
\centering
\small
\begin{tabular}{llc}
\toprule
\textbf{Protocol} & \textbf{Source / Split} & \textbf{\# Instances} \\
\midrule
\multirow{6}{*}{VBench-2.0 VQA}
  & Vinoground / Motion Order Understanding   & 201 \\
  & Vinoground / Dynamic Spatial Relationship & 35  \\
  & Vinoground / Camera-centric               & 64  \\
  & TimeBlind / Dynamic Attribute            & 51  \\
  & TimeBlind / Dynamic Spatial Relationship & 26  \\
\cmidrule{2-3}
  & \textit{Subtotal}                        & 377 \\
\midrule
Yes/No (direct) & TimeBlind / Unmapped         & 223 \\
\midrule
\multicolumn{2}{l}{\textbf{Total}}            & \textbf{600} \\
\bottomrule
\end{tabular}
\caption{Instance counts for the VQA-based evaluation of VBench-2.0. \textit{VBench-2.0 VQA} instances are evaluated via the three-question yes/no pipeline; \textit{Yes/No (direct)} instances use the benchmark's original binary questions without dimension mapping.}
\label{tab:temporal_protocol_counts}
\end{table*}

\section{Analysis of Reasoning Faithfulness and Human Validation}
\label{app:trace_analysis}
\begin{table}[t!]
\centering
\resizebox{\columnwidth}{!}{%
\begin{tabular}{@{}l l c@{}}
\toprule
\textbf{Stage} & \textbf{Metric} & \textbf{Score (\%)} \\
\midrule
\multirow{4}{*}{Plan Generation} 
  & Error-free                & 79.0 \\
  & Omission only             & 8.6  \\
  & Alteration only           & 9.6  \\
  & Both                      & 2.8  \\
\midrule
\multirow{4}{*}{Claim Verification} 
  & Evidence Grounding         & 80.9 \\
  & Claim Scope Alignment      & 93.6 \\
  & Judgment Validity          & 88.4 \\
  & Judgment Consistency       & 93.2 \\
\midrule
Score Aggregation 
  & Score Reproducibility             & 70.8 \\
\bottomrule
\end{tabular}%
}
\caption{
Reasoning trace diagnostics on VideoScore-Bench-v2, evaluated by GPT-5.4.
The table reports plan generation quality, claim verification faithfulness, and score aggregation consistency.
}
\label{tab:trace_analysis}
\end{table}

\subsection{Reasoning Trace Analysis}
\label{sec:reasoning-trace}
The ablations above show that the scene graph and plan-and-verify structure contribute to end-to-end performance. 
We further analyze whether the generated reasoning traces are faithful across three stages: whether the verification plan preserves the prompt semantics, whether claim-level verification is grounded and logically valid, and whether the final score is reproducible from the trace under the scoring rubric. 
We use GPT-5.4 as the evaluator on VideoScore-Bench-v2; Table~\ref{tab:trace_analysis} summarizes the results.

For plan generation, 79.0\% of verification plans are classified as error-free, while the remaining cases contain omissions (8.6\%), alterations (9.6\%), or both (2.8\%).
This indicates that most generated plans preserve the core visual semantics of the prompt.

Claim verification is assessed along four properties: whether rationales are grounded in the video or scene graph (Evidence Grounding), whether they stay within the scope of the corresponding claim (Claim Scope Alignment), whether the assigned judgment is supported by the rationale (Judgment Validity), and whether judgments are logically consistent across the trace (Judgment Consistency). 
The corresponding acceptance rates are 80.9\%, 93.6\%, 88.4\%, and 93.2\%, respectively. These results show that most claim-level rationales remain within the intended claim scope, support valid judgments, and maintain logical consistency across the trace, while also being largely grounded in the provided visual evidence.

% For score aggregation, we measure score reproducibility by asking GPT-5.4 to infer the final score from the model's reasoning trace, excluding the final analysis, using the same scoring rubric.
For score aggregation, we remove the final analysis and score and ask GPT-5.4 to reconstruct the final score from the preceding claim-level reasoning using the same rubric.
GPT-5.4 reproduces the model's final score in 70.8\% of cases. 
% The remaining 29.2\% reflects the role of the final analysis in score aggregation, which interprets claim judgments, claim importance, and grounded evidence under the rubric rather than mapping outcome counts to a fixed score.
The remaining 29.2\% indicates that the final score is not always reproducible from the preceding claim-level reasoning alone, motivating further analysis of the aggregation mechanism.

We further isolate two design choices in score aggregation: the rubric itself, and the Critical/Minor importance tags. 
Replacing the rubric with mechanical aggregation over the same per-claim outcomes—scalar mapping (Supported = 1, Partially Supported = 0.5, Contradicted = 0) followed by importance-weighted (Critical = 0.8, Minor = 0.2) or uniform averaging—drops VideoScore-Bench-v2 Alignment Accuracy from 39.8 to 35.0 and 36.2 respectively, a 3.6–4.8 point gap that confirms the model's score is not well captured by these simple fixed aggregation rules over outcome labels.

In contrast, forcing all importance tags to a single value (all-Critical or all-Minor) drops VSB-v2 alignment accuracy only to 39.0 and 38.2, indicating that the tags themselves contribute modestly. 
% The rubric-guided summarization, rather than the importance labeling, is the primary driver of correct score aggregation; the tags remain useful as an interpretability signal surfacing which requirements the model treats as core.
These results indicate that rubric-guided summarization contributes more to aggregation accuracy than the Critical/Minor labels themselves, while the labels remain useful as an interpretability signal indicating which requirements the model treats as core.

% In the preamble

\begin{table*}[t]
\centering
\small

\begin{tabular}{lcccc}
\toprule
\textbf{Evaluation criterion}
& \shortstack{\textbf{Human score}\\\textbf{(\%)}}
& \shortstack{\textbf{GPT score}\\\textbf{(\%)}}
& \shortstack{\textbf{Exact agreement}\\\textbf{(\%)}}
& \shortstack{\textbf{Within-one agreement}\\\textbf{(\%)}} \\
\midrule

\multicolumn{5}{l}{\textbf{Scene graph}} \\
\addlinespace[2pt]

Entity
& 85.21
& 100.00
& 53.61
& 82.22 \\

Attribute
& 84.17
& 99.22
& 49.58
& 90.00 \\

Relation
& 83.96
& 100.00
& 48.61
& 83.89 \\

\midrule

\multicolumn{5}{l}{\textbf{Reasoning trace}} \\
\addlinespace[2pt]

Plan Generation
& 86.25
& 93.75
& 75.28
& -- \\

Evidence Grounding
& 89.10
& 84.90
& 89.10
& -- \\

Claim-Scope Alignment
& 96.03
& 96.30
& 92.88
& -- \\

Judgment Validity
& 90.27
& 96.22
& 89.18
& -- \\

Judgment Consistency
& 90.00
& 100.00
& 83.61
& -- \\

\bottomrule
\end{tabular}

\caption{
Human evaluation of scene-graph quality and reasoning-trace faithfulness,
with corresponding GPT scores and human inter-annotator agreement.
}
\label{tab:human_eval}

\end{table*}
\subsection{Human Evaluation}

\paragraph{Evaluation Setup.}
To further validate the reliability of our automated evaluation, we conduct a targeted human evaluation of SG-PVR's scene graphs and reasoning traces. We sample 16 outputs, consisting of eight examples each from Temporal Semantics and VideoScore-Bench-v2. 
To cover both typical model behavior and outputs rated highly by the automated evaluator, the evaluation set includes eight randomly selected outputs and eight outputs that received perfect automated scores for both scene-graph quality and reasoning-trace faithfulness.

Ten human evaluators participate in the study.
All ten evaluators independently assess all 16 outputs.
Each evaluator is provided with detailed metric definitions, scoring rubrics, and annotated examples, while the automated GPT evaluation results are withheld to avoid bias. 
For scene-graph quality, evaluators assess the correctness of entities, attributes, and relations on a 0--3 ordinal scale, and we report the mean rating normalized to a percentage.
For reasoning-trace faithfulness, we evaluate Plan Generation and the same four claim-verification criteria used in Appendix~\ref{sec:reasoning-trace}, with each evaluation unit labeled as valid or invalid at the claim or sample level, depending on the criterion.
Inter-annotator agreement is measured by exact agreement on these labels; for the ordinal scene-graph ratings, we additionally report within-one agreement, which counts ratings differing by at most one point.

\paragraph{Annotator Recruitment and Consent.}
The human annotators involved in the Vinoground filtering described in Appendix~\ref{app:temporal_details} and the human evaluation above were members of our research group who volunteered to participate. 
No monetary compensation was provided. 
Before annotation, all participants were informed that their judgments would be used for research purposes and reported in this paper, and they agreed to participate. 
We did not collect demographic or other personally identifying information from the annotators.

\paragraph{Human Evaluation Instructions.}
The instructions were originally presented to the evaluators in Korean through the annotation interface.
Table~\ref{tab:human_eval_instructions} provides an English translation.

\paragraph{Results.}
As shown in Table~\ref{tab:human_eval}, human evaluators assign consistently high scores to both components of SG-PVR. Scene-graph quality ranges from 83.96\% to 85.21\%, while reasoning-trace scores range from 86.25\% to 96.03\%. Inter-annotator agreement is also high: within-one agreement reaches 82.22--90.00\% for the ordinal scene-graph ratings, and exact agreement ranges from 75.28\% to 92.88\% across the reasoning-trace criteria. These results provide direct human evidence that the generated scene graphs and intermediate reasoning traces are generally faithful and consistently judged across annotators.

Human and automated evaluations also show broadly similar trends for reasoning-trace faithfulness. 
In particular, Evidence Grounding and Claim-Scope Alignment receive closely matched human and GPT scores (89.10\% vs. 84.90\%, and 96.03\% vs. 96.30\%, respectively). 
For scene-graph quality, the automated evaluator tends to assign more optimistic scores than human annotators; nevertheless, human ratings remain consistently above 83\% across entities, attributes, and relations. 
% Overall, the human evaluation supports both the quality of SG-PVR's intermediate representations and the reliability of our automated evaluation for analyzing reasoning-trace faithfulness.
Overall, the human evaluation provides complementary evidence for the quality of SG-PVR’s intermediate representations and broadly supports the trends observed in our automated reasoning-trace analysis.

\section{Additional Analyses}
\label{app:additional_analysis}
\subsection{Failure Mode Analysis}
\label{sec:failure-mode}
% In the preamble

\begin{table*}[t]
\centering
\small
\resizebox{\textwidth}{!}{%
\begin{tabular}{@{}c c c c c c c c c c@{}}
\toprule
\textbf{$d$}
& \textbf{$n$}
& \shortstack{\textbf{SG Precision}\\\textbf{(E/A/R) (\%)}}
& \shortstack{\textbf{SG Coverage}\\\textbf{Avg. (\%)}}
& \shortstack{\textbf{Plan Generation}\\\textbf{(\%)}}
& \shortstack{\textbf{Evidence Grounding}\\\textbf{(\%)}}
& \shortstack{\textbf{Claim-Scope Alignment}\\\textbf{(\%)}}
& \shortstack{\textbf{Judgment Validity}\\\textbf{(\%)}}
& \shortstack{\textbf{Judgment Consistency}\\\textbf{(\%)}}
& \shortstack{\textbf{Score Reproducibility}\\\textbf{(\%)}} \\
\midrule

$1$
& 214
& 99.07 / 90.67 / 84.96
& 33.81
& 76.17
& 80.18
& 93.40
& 88.01
& 94.39
& 69.63 \\

$\geq 2$
& 87
& 99.71 / 93.67 / 89.02
& 34.38
& 81.61
& 82.20
& 93.43
& 89.59
& 91.95
& 63.22 \\

\bottomrule
\end{tabular}%
}

\caption{
Pipeline-stage failure analysis on VideoScore-Bench-v2, grouped by absolute
score error \(d = |\mathrm{GT}-\mathrm{Pred}|\).
We compare scene-graph quality and reasoning-trace faithfulness between
examples with \(d=1\) and \(d\geq2\).
}
\label{tab:failure-mode}

\end{table*}

\label{sec:failure-mode}

% To investigate where large prediction errors arise in the SG-PVR pipeline, we conduct a failure analysis across its major pipeline stages on VideoScore-Bench-v2. 
% We consider the 301 examples for which the predicted score differs from the ground-truth (GT) score and group them according to the absolute score error \(d = |\mathrm{GT}-\mathrm{Pred}|\): 214 examples with \(d=1\) and 87 examples with \(d\geq2\). We compare the two groups across the major stages of SG-PVR: Scene-Graph Generation, Plan Generation, Claim Verification, and Score Aggregation. Scene-graph quality is measured using entity, attribute, and relation precision together with prompt-relevant SG coverage, while the remaining stages are evaluated using the reasoning-trace criteria described in Appendix~\ref{sec:reasoning-trace}. Results are summarized in Table~\ref{tab:failure-mode}.

To characterize which pipeline diagnostics are associated with larger prediction errors, we analyze the 301 VideoScore-Bench-v2 examples for which the predicted score differs from the ground-truth (GT) score. 
We group them by absolute score error \(d = |\mathrm{GT}-\mathrm{Pred}|\): 214 examples with \(d=1\) and 87 with \(d\ge2\). 
We compare the two groups across Scene-Graph Generation, Plan Generation, Claim Verification, and Score Aggregation. 
Scene-graph quality is measured using entity, attribute, and relation precision together with prompt-relevant SG coverage, while the remaining stages use the reasoning-trace criteria described in Appendix~\ref{sec:reasoning-trace}. 
Results are summarized in Table~\ref{tab:failure-mode}.

% We first examine whether larger score errors are associated with lower scene-graph quality. Entity, attribute, and relation precisions are 99.07\%, 90.67\%, and 84.96\%, respectively, for \(d=1\), and slightly increase to 99.71\%, 93.67\%, and 89.02\% for \(d\geq2\). Prompt-relevant SG coverage also remains nearly unchanged, from 33.81\% to 34.38\%. Thus, within incorrectly scored examples, larger score discrepancies are not associated with lower scene-graph precision or coverage.

We first examine scene-graph quality. 
Entity, attribute, and relation precisions are 99.07\%, 90.67\%, and 84.96\%, respectively, for \(d=1\), compared with 99.71\%, 93.67\%, and 89.02\% for \(d\ge2\). 
Prompt-relevant SG coverage also remains nearly unchanged, from 33.81\% to 34.38\%. 
Thus, within incorrectly scored examples, larger prediction errors are not associated with lower scene-graph precision or coverage.

% Plan Generation and Claim Verification similarly show little degradation as the score error increases. The error-free plan-generation rate increases from 76.17\% for \(d=1\) to 81.61\% for \(d\geq2\). Evidence Grounding increases from 80.18\% to 82.20\%, Claim-Scope Alignment remains nearly identical (93.40\% vs.\ 93.43\%), and Judgment Validity increases from 88.01\% to 89.59\%. Judgment Consistency decreases modestly from 94.39\% to 91.95\%, but we observe no systematic deterioration across the plan-generation or claim-verification criteria for examples with larger score errors. This suggests that large prediction errors are not primarily characterized by failures in constructing the verification plan or verifying individual claims.

Plan Generation and Claim Verification show a similar pattern. The error-free plan-generation rate changes from 76.17\% for \(d=1\) to 81.61\% for \(d\ge2\). 
Evidence Grounding changes from 80.18\% to 82.20\%, Claim-Scope Alignment remains nearly identical (93.4\% vs. 93.43\%), and Judgment Validity changes from 88.01\% to 89.59\%. 
Judgment Consistency decreases modestly from 94.39\% to 91.95\%. 
Overall, these diagnostics do not show systematic degradation as the score error increases.

% The clearest degradation appears at the Score Aggregation stage. Score Reproducibility decreases from 69.63\% for \(d=1\) to 63.22\% for \(d\geq2\), a drop of 6.41 percentage points. Score Reproducibility measures whether the final scalar score can be reconstructed from the preceding claim-level judgments under the same scoring rubric. The lower reproducibility among large-error examples therefore indicates a stronger association between large prediction errors and instability in translating claim-level verification outcomes into the final scalar score.

Among the measured diagnostics, the largest decrease occurs in Score Reproducibility, which drops from 69.63\% for \(d=1\) to 63.22\% for \(d\ge2\), a difference of 6.41 percentage points. 
Score Reproducibility measures whether the final scalar score can be reconstructed from the preceding claim-level reasoning under the same scoring rubric. 
This descriptive difference suggests that larger prediction errors are more often accompanied by inconsistency between the preceding reasoning and the final aggregated score.

% Overall, we do not observe increasing failure rates in Scene-Graph Generation, Plan Generation, or Claim Verification as the prediction discrepancy grows. Instead, the most pronounced difference between small- and large-error cases occurs at the final Score Aggregation stage. 
Overall, Scene-Graph Generation, Plan Generation, and Claim Verification show no systematic degradation with increasing error magnitude, whereas Score Reproducibility exhibits the largest observed difference between the two groups. 
% These results suggest that improving the consistency of rubric-guided score aggregation may be particularly important for reducing large scoring errors, while also indicating that such errors cannot be explained simply by degradation in scene-graph quality or individual claim verification.
These results identify score aggregation as the stage showing the largest diagnostic difference between small- and large-error cases, motivating more consistent aggregation as a direction for improvement.

\subsection{Effect of Base Model Choice}
\label{app:base_model}

\begin{table}[t!]
\centering
\resizebox{\columnwidth}{!}{
\begin{tabular}{llcc}
\toprule
\textbf{Method} & \textbf{Base Model} & \makecell{\textbf{Temp. Overall}\\\textbf{(Task1 / Task2)}} & \textbf{MJ-Bench-Video} \\
\midrule
VideoScore2 (SFT Only) & Qwen2.5-VL-7B & 10.5 (35.5 / 34.8) & 64.0 \\ 
VideoScore2 (SFT Only) & Qwen3.5-9B & 11.9 (35.6 / 35.4) & 63.6 \\ 
SG-PVR & Qwen3.5-9B & \textbf{31.4 (51.3 / 51.9)} & \textbf{68.2} \\ 
\bottomrule 
\end{tabular} } 
\caption{Effect of base model choice.} 
\label{tab:base_model} 
\end{table}

To assess whether SG-PVR's gains can be explained by its stronger Qwen3.5-9B base model, we train a Qwen3.5-9B variant following the SFT stage of VideoScore2, using its original training data and configuration.
We compare this variant with the released VideoScore2 SFT-only checkpoint based on Qwen2.5-VL-7B and with SG-PVR.
As shown in Table~\ref{tab:base_model}, replacing Qwen2.5-VL-7B with Qwen3.5-9B under the same VideoScore2 SFT setting does not improve MJ-Bench-Video accuracy and changes Temporal Overall only marginally from 10.5 to 11.9.
In contrast, SG-PVR with the same Qwen3.5-9B base model achieves 68.2 on MJ-Bench-Video and 31.4 Temporal Overall.
These results indicate that SG-PVR's gains cannot be explained by the base model alone.

Table~\ref{tab:ablation} further evaluates SG-PVR ablations using the same Qwen3.5-9B base model and comparably sized supervision constructed from the same source datasets.
Replacing plan-and-verify with free-form reasoning reduces Temporal Overall from 31.4 to 21.8 with video and scene-graph grounding, and from 30.5 to 25.6 under video-only grounding.
Together, these results support the contribution of structured plan-and-verify reasoning to fine-grained temporal semantic evaluation.

% \paragraph{Effect of direct teacher supervision.}
% We further test whether the gains can be attributed simply to supervision from the stronger Qwen3.5-27B teacher.
% Using the same Stage-2 source samples, scene graphs, and data scale as SG-PVR, we train a teacher-score variant with the same SG-grounded plan-and-verify output structure.
% Unlike SG-PVR, which retains or regenerates teacher traces under GT-score matching and consistency filtering, this variant directly uses the teacher-generated reasoning and its predicted score as the supervision target, without GT-score anchoring.
% It obtains 23.2/21.2/30.4 VSB-v2 Accuracy (V/P/A) and 27.5 Temporal Overall, compared with 43.6/38.2/39.8 and 31.4 for SG-PVR.
% This substantial gap shows that directly distilling the teacher's own reasoning and reward predictions is insufficient to explain SG-PVR's performance.
% Together with the same-teacher ablations in Table~\ref{tab:ablation}, these results indicate that the gains are not explained by teacher distillation alone.

\subsection{Additional Reranking Analysis}
\label{app:semantic_reranking}

\begin{table}[t!]
\centering
\resizebox{\columnwidth}{!}{%
\begin{tabular}{@{}llccccccc@{}}
\toprule
\textbf{Generator} & \textbf{Method} & \textbf{C-Attr} & \textbf{D-Attr} & \textbf{Spat.} & \textbf{Mot.} & \textbf{Act.} & \textbf{Inter.} & \textbf{Num.} \\ \midrule
Wan2.1-1.3B  & Random          & 77.5          & 2.7          & 53.4          & 38.3          & 61.9          & 59.5          & 45.7          \\
             & OmniScore       & {\ul 82.1}    & 3.2          & 54.8          & {\ul 39.0}    & \textbf{72.1} & \textbf{68.7} & 51.2          \\
             & VideoScore2     & 80.2          & {\ul 3.3}    & {\ul 55.1}    & 38.8          & 68.5          & 61.7          & {\ul 56.9}    \\ \cmidrule(l){2-9} 
             & \textbf{SG-PVR} & \textbf{82.5} & \textbf{4.2} & \textbf{58.3} & \textbf{39.1} & {\ul 69.6}    & {\ul 63.5}    & \textbf{59.6} \\ \midrule
CogVideoX-2B & Random          & 77.9          & 1.4          & 50.0          & 38.3          & 48.4          & 59.2          & 36.4          \\
             & OmniScore       & {\ul 83.1}    & 1.1          & {\ul 56.8}    & \textbf{39.0} & \textbf{61.6} & \textbf{65.7} & 41.9          \\
             & VideoScore2     & 82.0          & {\ul 1.5}    & 53.8          & {\ul 38.9}    & 58.8          & 60.9          & {\ul 42.7}    \\ \cmidrule(l){2-9} 
             & \textbf{SG-PVR} & \textbf{83.6} & \textbf{2.2} & \textbf{61.1} & {\ul 38.9}    & {\ul 59.5}    & {\ul 62.1}    & \textbf{48.1} \\ \bottomrule
\end{tabular}
}
\caption{Test-time reranking on T2V-CompBench using semantic-only reward signals.
For all dimensions, higher is better.}
\label{tab:semantic_reranking}
\end{table}

The main reranking experiment uses the aggregated reward signal of each model.
To isolate semantic alignment from video quality, we repeat the experiment using only the semantic-alignment score of each reward model.
For SG-PVR, we use only the SA score, excluding VQ, TQ, and PC.
For VideoScore2, we use Text Alignment score, and for OmniScore, we use Overall Consistency, its ViCLIP-based text-video semantic alignment score.
The candidate pools and evaluation protocol are otherwise identical to the main reranking experiment.

As shown in Table~\ref{tab:semantic_reranking}, SG-PVR outperforms VideoScore2 on all seven dimensions for Wan2.1-1.3B and on six dimensions with one tie for CogVideoX-2B.
Compared with OmniScore, SG-PVR achieves higher scores on five of seven dimensions for Wan2.1-1.3B and four of seven dimensions for CogVideoX-2B.
SG-PVR consistently achieves the best results on Consistent Attribute, Dynamic Attribute, Spatial Relationship, and Numeracy across both generators.
These results show that the reranking gains persist when only semantic-alignment scores are used, indicating that the downstream improvement is not driven by SG-PVR's additional quality scores.

\subsection{Generalization Analysis}
\label{app:generalization}

\paragraph{Temporal Complexity and Video Duration.}

\begin{table}[t!]
\centering
\resizebox{\columnwidth}{!}{
\begin{tabular}{lcccc}
\toprule
\makecell{\textbf{Event}\\ \textbf{Count}}
& \makecell{\textbf{Avg.}\\ \textbf{Duration (s)}}
& \makecell{\textbf{SG Precision}\\ \textbf{E / A / R}}
& \makecell{\textbf{SG Coverage}\\ \textbf{E / A / Act.}}
& \makecell{\textbf{Temp.}\\ \textbf{Overall}} \\
\midrule
1
& 4.77
& 99 / 90 / 87
& 69 / 17 / 27
& 35.14 \\
$\geq 2$
& 11.70
& 99 / 90 / 80
& 60 / 21 / 25
& 32.67 \\
\bottomrule
\end{tabular}
}
\caption{
Scene-graph quality and reward accuracy across temporal complexity levels.
Each subset contains 150 examples, and all scene-graph metrics are reported in percentage.
}
\label{tab:duration_complexity}
\end{table}

We examine the robustness of SG-PVR as temporal complexity and video duration increase jointly.
Following the event-count analysis in \S\ref{sec:finegrained_analysis}, we construct two equally sized subsets of 150 examples from the temporal semantics test set.
The single-event subset has an average duration of 4.77 seconds, while the multi-event subset contains at least two events and has an average duration of 11.70 seconds.
As temporal complexity and average duration increase, entity and attribute precision remain stable at approximately 99\% and 90\%, while relation precision changes moderately from 87\% to 80\%.
Scene-graph coverage remains broadly comparable across the two subsets.
SG-PVR accuracy also remains relatively stable, changing from 35.14\% to 32.67\% despite the increase in both event count and average video duration.
Overall, these results show that SG-PVR maintains comparable scene-graph quality and reward accuracy under increased temporal complexity and duration.

\paragraph{Unseen Generator.}

\begin{table}[t!]
\centering
\resizebox{\columnwidth}{!}{
\begin{tabular}{lccc}
\toprule
\textbf{Generator} & \textbf{Seen} & \textbf{SG-PVR} & \textbf{VideoScore2 (SFT Only)} \\
\midrule
Gen-2           & \ding{51} & 51.8 / 45.6 & 37.8 / 34.9 \\
Hotshot-XL      & \ding{51} & 41.9 / 32.7 & 36.8 / 31.4 \\
VideoFusion     & \ding{51} & 53.6 / 48.7 & 36.4 / 33.4 \\
VideoCrafter    & \ding{51} & 43.4 / 40.0 & 30.4 / 27.9 \\
Text2Video-Zero & \ding{51} & 55.8 / 52.3 & 39.2 / 34.2 \\
Tune-A-Video    & \ding{55} & 42.6 / 39.9 & 41.1 / 41.1 \\
\bottomrule
\end{tabular}}
\caption{Per-generator alignment correlation on LGVQ, reported as PLCC / SRCC.
Seen indicates whether the generator is represented in SG-PVR's Stage-2 training sources.}
\label{tab:generator_generalization}
\end{table}

Our Stage-2 training sources cover more than 35 T2V generators spanning diverse generation paradigms, including U-Net diffusion, diffusion transformers, autoregressive models, and commercial systems. 
To evaluate generalization beyond these training generators, we conduct a per-generator analysis on LGVQ, which provides generator labels for all videos. 
Among its six generators, Tune-A-Video does not appear in the training data of either SG-PVR or VideoScore2 (SFT Only), providing an unseen-generator subset for both models. 
Table~\ref{tab:generator_generalization} reports Text-Video Alignment PLCC and SRCC separately for each generator. 
On Tune-A-Video, SG-PVR achieves 42.6 PLCC and 39.9 SRCC, compared with 41.1 PLCC and 41.1 SRCC for VideoScore2 (SFT Only). 
SG-PVR's PLCC on this unseen generator also falls within the range observed across the five generators represented in its training sources, which ranges from 41.9 to 55.8. 
Moreover, SG-PVR outperforms VideoScore2 (SFT Only) on all five seen-generator subsets. 
These results show no evident degradation associated with the unseen generator and suggest that SG-PVR's alignment performance is not restricted to generators represented during training.

\section{System Prompts}
\label{app:systemprompts}
\subsection{System Prompts for Reasoning Trace Construction}
\label{app:system_prompts}

\subsubsection{Verification Plan Generation}

\begin{tcolorbox}[
  enhanced jigsaw,
  breakable,
  pad at break*=1mm,
  colback=blue!3!white,
  colframe=blue!65!black,
  title={System Prompt for Verification Plan Generation},
  fonttitle=\bfseries,
  coltitle=white,
  boxrule=0.5pt,
  arc=1mm,
  left=5pt,
  right=5pt,
  top=5pt,
  bottom=5pt
]
\footnotesize

You are an expert planner for prompt-to-video semantic alignment evaluation.

Your task is to read a text-to-video Prompt and produce a concise Verification Plan.

\vspace{0.4em}
\noindent\textbf{[Definition]}

\noindent- Criterion: a semantic evaluation category.

\noindent- Verification Claim: a short atomic statement describing one thing to verify.

\vspace{0.4em}
\noindent\textbf{[Task]}

\noindent 1. Identify only the Criteria that are necessary for evaluating the Prompt.

\noindent 2. Decompose the Prompt into atomic Verification Claims under those Criteria.

\noindent 3. Assign exactly one semantic importance label to each claim: Critical or Minor

\noindent \hspace*{1em}- Critical: main subjects, primary actions/events, key attributes, key spatial relations, temporal order, or causal/event structure.

\noindent \hspace*{1em}- Minor: background details, secondary objects, lighting, style, camera angle, or other non-essential visual details.

\vspace{0.4em}
\noindent\textbf{[Criteria]}

\noindent- Entity: subjects, objects, or characters explicitly mentioned or clearly implied.

\noindent- Attribute: properties or states of an entity.

\noindent- Action: motions, activities, or behaviors.

\noindent- Spatial Relation: spatial arrangement or positioning between entities.

\noindent- Temporal Constraint: temporal order, transition, or dependency between actions or events.

\vspace{0.4em}
\noindent\textbf{[Rules]}

\noindent- Include only Criteria that are required by the Prompt.

\noindent- Omit any Criterion with no relevant content.

\noindent- Use only information explicitly stated or clearly implied in the Prompt.

\noindent- Do not add unspecified details.

\noindent- Do not include video quality, visual quality, style, or aesthetics.

\noindent- Each Verification Claim must begin with ``Verify whether ...'', be atomic, be concise, not overlap with other claims, and not repeat the same semantic content under different Criteria.

\noindent- Merge redundant or near-duplicate claims.

\noindent- Prefer the minimum number of claims needed to cover the Prompt's core semantic requirements.

\noindent- Keep the overall plan concise.

\vspace{0.4em}
\noindent\textbf{[Output Format]}

\noindent Output only plain text in the following format:

\vspace{0.4em}
\noindent 1. [Criterion Name]\\
\noindent- Verify whether ... (Critical)\\
\noindent- Verify whether ... (Minor)

\vspace{0.4em}
\noindent 2. [Criterion Name]\\
\noindent- Verify whether ... (Critical)

\end{tcolorbox}
\subsubsection{Semantic Reasoning Trace Generation.}
\label{app:semantic_reasoning_trace_generation}
\begin{tcolorbox}[
  enhanced jigsaw,
  breakable,
  pad at break*=1mm,
  colback=blue!3!white,
  colframe=blue!65!black,
  title={System Prompt for Semantic Alignment Reasoning Generation},
  fonttitle=\bfseries,
  coltitle=white,
  boxrule=0.5pt,
  arc=1mm,
  left=5pt,
  right=5pt,
  top=5pt,
  bottom=5pt
]
\footnotesize

You are an expert Video Semantic Alignment Evaluator.

Your task is to evaluate whether the video and video scene graph satisfy the semantic requirements of the original Prompt by strictly following the Verification Plan.

\vspace{0.4em}
\noindent\textbf{[Input]}

\noindent You will be given:

\noindent 1. the original generation prompt

\noindent 2. a Verification Plan

\noindent 3. video

\noindent 4. a video scene graph

\noindent The video scene graph consists of an object scene graph and a relationship scene graph.

\noindent Each subject/object ID in the relationship scene graph refers to the corresponding object ID in the object scene graph. If object\_id is -1, it indicates None.

\vspace{0.4em}
\noindent\textbf{[Tasks]}

\noindent Before writing, carefully inspect the video and video scene graph in full. Do not rush.

\noindent Evaluate every claim in the Verification Plan in the exact order it appears.

\noindent For each criterion heading in the Verification Plan, keep the same numbered heading.

\noindent Under each heading, output one bullet point per claim.

\noindent Each bullet point must correspond to the claim in the same order under that criterion.

\vspace{0.4em}
\noindent Perform a step-by-step evaluation for each claim using the following process:

\noindent 1. Identify what the claim requires.

\noindent 2. Inspect the video and video scene graph and find specific verifiable evidence related to the claim.

\noindent 3. Decide whether the evidence supports, partially supports, or contradicts the claim.

\noindent 4. Write one concise bullet point that first states the specific evidence and briefly explains why it justifies the judgment, then ends with the Judgment label in parentheses.

\vspace{0.4em}
\noindent After evaluating all claims, write a concise Final Analysis explaining the final score based on the distribution of Critical/Minor claims and their judgments.

\noindent Finally, assign one final score from 1 to 5 formatted as \verb|\boxed{Score}|.

\vspace{0.4em}
\noindent\textbf{[Rules]}

\noindent 1. Do not rewrite, quote, or summarize the claim itself.

\noindent 2. Evaluate each claim independently. Do not merge multiple claims into one bullet.

\noindent 3. Use the exact criterion headings from the Verification Plan and preserve their order.

\noindent 4. Evidence must be grounded in the video and video scene graph. State what is actually observed or stated, not what should be present.

\noindent 5. Evidence from either the video or the scene graph is sufficient to support a claim.

\noindent 6. If an element required by the claim is completely missing from the video and video scene graph, label the judgment as Contradicted.

\noindent 7. If the main requirement is present but a secondary detail is missing, incomplete, or ambiguous, label the judgment as Partially Supported.

\noindent 8. Each evaluation bullet must end with exactly one judgment label in this parentheses: (Supported), (Partially Supported), or (Contradicted).

\noindent 9. Do not use field headers such as ``Evidence:'', ``Reasoning:'', or ``Judgment:'' inside the evaluation bullets.

\noindent 10. Output ONLY the specified format.

\vspace{0.4em}
\noindent\textbf{[Score Definition]}

\noindent - 5 (Excellent): ALL claims, both Critical and Minor, are Supported.

\noindent - 4 (Good): All Critical claims are Supported, or only 1--2 Critical claims are Partially Supported. At most 1--2 Minor claims are Contradicted.

\noindent - 3 (Fair): 1--2 Critical claims are Contradicted, such as a missing specific event or object, but the core meaning remains recognizable.

\noindent - 2 (Poor): 3 or more Critical claims are Contradicted, such as a broken core temporal sequence, or only 1--2 Critical claims are Supported.

\noindent - 1 (Fail): ALL Critical claims are Contradicted. Any Supported or Partially Supported claims are strictly Minor.

\vspace{0.4em}
\noindent\textbf{[Output Format]}

\noindent Output ONLY in the following format.

\vspace{0.4em}
\noindent 1. [Criterion Name in Verification Plan]\\
\noindent - [Specific evidence from the video and video scene graph and brief reasoning explaining why the evidence justifies the judgment.] [Supported / Partially Supported / Contradicted]\\
\noindent - [Specific evidence from the video and video scene graph and brief reasoning explaining why the evidence justifies the judgment.] [Supported / Partially Supported / Contradicted]

\vspace{0.4em}
\noindent 2. [Criterion Name in Verification Plan]\\
\noindent - [Specific evidence from the video and video scene graph and brief reasoning explaining why the evidence justifies the judgment.] [Supported / Partially Supported / Contradicted]\\
\noindent - [Specific evidence from the video and video scene graph and brief reasoning explaining why the evidence justifies the judgment.] [Supported / Partially Supported / Contradicted]

\vspace{0.4em}
\noindent ... [Continue until all criteria in the Verification Plan have been evaluated.]

\vspace{0.4em}
\noindent Final Analysis:

\noindent [Concise explanation of the final score based on the distribution of Critical/Minor claims and their judgments.]

\noindent Score: \verb|\boxed{Score}|

\end{tcolorbox}

\subsubsection{Quality Reasoning Trace Generation.}
\label{app:quality_reasoning_trace_generation}
\begin{tcolorbox}[
  enhanced jigsaw,
  breakable,
  pad at break*=1mm,
  colback=blue!3!white,
  colframe=blue!65!black,
  title={System Prompt for Quality Reasoning Generation},
  fonttitle=\bfseries,
  coltitle=white,
  boxrule=0.5pt,
  arc=1mm,
  left=5pt,
  right=5pt,
  top=5pt,
  bottom=5pt
]
\footnotesize

You are an expert evaluator of AI-generated videos.

Evaluate the video across three quality dimensions independently. Each dimension requires different expertise and focus.

\vspace{0.4em}
\noindent\textbf{[Dimension 1: Visual Quality]}

\noindent Per-frame optical and perceptual quality. Ignore motion and content --- judge what each frame looks like as a still image.

\noindent Check: noise/grain, blur, exposure, compression artifacts, color cast, resolution, sharpness of main subject vs.\ background, banding, ringing.

\vspace{0.4em}
\noindent\textbf{[Dimension 2: Temporal Quality]}

\noindent Frame-to-frame stability and coherence. About stability, NOT amount of motion.

\noindent Check: object identity persistence across frames (no morphing into a different object), anatomical persistence (consistent finger/limb count), flickering not explained by scene physics, motion smoothness (no teleporting/stuttering), background stability of static elements.

\noindent Intentional scene cuts and natural motion blur are not defects.

\vspace{0.4em}
\noindent\textbf{[Dimension 3: Physical / Common-sense Consistency]}

\noindent Whether the depicted world obeys real-world physics and common sense.

\noindent Check: gravity, solid-body physics and collisions, anatomy, spatial relations (perspective, occlusion, scale), shadows/reflections vs.\ light source, fluid/particle physics, causality, contextual common sense.

\noindent Stylized or fantastical content is judged by its own internal consistency.

\vspace{0.4em}
\noindent\textbf{[Analysis Requirements]}

\noindent- For each dimension, write 3 to 5 sentences of detailed, professional analysis grounded in direct visual evidence.

\noindent- For every defect you mention, ground it in three ways: (i) WHERE in space --- a specific region (e.g., ``the subject's left hand'', ``the background wall''); (ii) WHEN --- the timestamp or part of the clip where it occurs; and (iii) HOW MUCH --- whether it happens once, intermittently, or persists, and roughly how much of the clip it spans. The timestamp alone is not enough: always state the persistence/extent, because that is what determines severity. Generic claims detached from this video (e.g., ``typical AI artifacts'') are not acceptable.

\noindent- For high-quality dimensions, describe the positive evidence (stable identities, consistent textures, plausible interactions), not merely the absence of defects.

\noindent- Judge each dimension independently --- a defect in one dimension must not affect the score of another.

\noindent- Do NOT reference rubrics, tiers, or numeric scores within the analysis text. The analysis must read as a standalone expert assessment.

\noindent- Describe defects by what they concretely are, where they occur, and how often --- not by reusing the rubric's band phrases (e.g., do not write ``noticeable during normal playback'' or ``minor issue on close inspection'' as a label). State the observable fact; let the severity be inferable from it.

\noindent- The length of the analysis must NOT correlate with the score.

\vspace{0.4em}
\noindent\textbf{[Scoring Scale (per dimension)]}

\noindent- 5: no observable issues.

\noindent- 4: minor issues, visible only on close inspection.

\noindent- 3: noticeable issues during normal playback, generally watchable.

\noindent- 2: frequent or severe issues that dominate the viewing experience.

\noindent- 1: pervasive failures; the dimension is severely degraded.

\vspace{0.4em}
\noindent\textbf{[Output Format]}

\noindent Respond in EXACTLY this format. No markdown, no code fences, no headings, no text outside the tags. Do NOT emit JSON.

\vspace{0.4em}
\noindent \verb|<think>|\\
\noindent 1. Visual Quality: \verb|<3-5 sentence analysis>| Score: \verb|<1-5>|\\
\noindent 2. Temporal Quality: \verb|<3-5 sentence analysis>| Score: \verb|<1-5>|\\
\noindent 3. Physical/Common-sense Consistency: \verb|<3-5 sentence analysis>| Score: \verb|<1-5>|\\
\noindent \verb|</think>|\\
\noindent \verb|<answer>VQ <1-5> TQ <1-5> PC <1-5></answer>|

\vspace{0.4em}
\noindent\textbf{[Format Constraints]}

\noindent- The \verb|<answer>| block MUST contain the same scores as the \verb|<think>| block.

\noindent- The \verb|<answer>| block uses the EXACT pattern \verb|VQ <n> TQ <n> PC <n>| --- uppercase, single spaces, no commas, no extra punctuation, no newlines, on one line.

\noindent- Each \verb|Score: <n>| appears at the END of its dimension's paragraph in \verb|<think>|.

\noindent- Open \verb|<think>| once and close once before \verb|<answer>|. Open \verb|<answer>| once and close once at the end.

\noindent- Output nothing before \verb|<think>| and nothing after \verb|</answer>|.

\end{tcolorbox}

\subsection{System Prompts for SG-PVR Inference}
\label{app:system_prompts_inference}

\begin{tcolorbox}[
  enhanced jigsaw,
  breakable,
  pad at break*=1mm,
  colback=blue!3!white,
  colframe=blue!65!black,
  title={System Prompts for SG-PVR Inference},
  fonttitle=\bfseries,
  coltitle=white,
  boxrule=0.5pt,
  arc=1mm,
  left=5pt,
  right=5pt,
  top=5pt,
  bottom=5pt
]
\footnotesize

You are an expert video reward model. Given a text prompt and a video generated from that prompt, evaluate it along two axes: (1) semantic alignment with the prompt, and (2) intrinsic video quality. 

\vspace{0.4em}
\noindent Reason by first constructing a scene graph from the video, then using it as intermediate evidence for semantic verification.

\noindent You must follow the four steps below in order and produce the output in the exact structure specified. Steps 1--3 are performed inside \verb|<think>...</think>|; Step 4 emits the final answer after closing \verb|</think>|. Do not skip steps, reorder them, or omit any tag.

\vspace{0.4em}
\noindent\textbf{[STEP 1 - Scene Graph]}

\noindent Observe the video and emit a spatio-temporal scene graph capturing the key entities, their attributes, and the relationships or events occurring over time. Wrap the entire scene graph in \verb|<scene_graph>...</scene_graph>|.

\vspace{0.4em}
\noindent\textbf{Tasks:}

\noindent 1. Identify the key entities with their categories and attributes.

\noindent 2. Identify the important relationships and events with temporal ranges.

\noindent 3. Emit the result as a single JSON object inside \verb|<scene_graph>| with:

\noindent \texttt{\{}

\noindent \texttt{~~~~"objects": [\{"id": "<id>", "category": "<noun>", "attributes": ["<adj>", ...]\}],}

\noindent \texttt{~~~~"relationships": [["<subj\_id>", "<predicate>", "<obj\_id>", [[<start>, <end>], ...], "<type>"], ...]}

\noindent \texttt{\}}

\vspace{0.4em}
\noindent\textbf{Rules:}

\noindent - Object IDs are unique strings starting from "0". Use "-1" to represent the camera.

\noindent - Relations must have temporal ranges. Time values are in seconds from video start.

\noindent - Focus on the visually and semantically important entities and relationships.

\noindent - Do not invent attributes or actions that are not visible.

\vspace{0.4em}
\noindent\textbf{[STEP 2 - Semantic Alignment]}

\noindent Evaluate how faithfully the video realizes the semantic requirements of the prompt, using the scene graph as explicit intermediate evidence. Wrap the entire output in \verb|<semantic>...</semantic>|.

\vspace{0.4em}
\noindent\textbf{Tasks:}

\noindent 1. In \verb|<plan>...</plan>|, decompose the prompt into atomic Verification Claims covering explicit and clearly implied semantic requirements: Entity, Attribute, Action, Spatial Relation, and Temporal Constraints. For each claim, assign a semantic importance label: (Critical/Minor)

\noindent 2. Using the video and scene graph as evidence, evaluate each claim in the Verification Plan in order. For each claim, write one bullet with specific evidence and brief reasoning, ending with the Judgment label in parentheses: (Supported/Partially Supported/Contradicted).

\noindent 3. After evaluating all claims, write `Final Analysis:' to summarize the distribution of claim judgments, while considering the semantic importance labels assigned in \verb|<plan>|, then end with \verb|Semantic Score: <1-5>|.

\vspace{0.4em}
\noindent\textbf{Rules:}

\noindent - Do not introduce or evaluate claims outside \verb|<plan>|.

\noindent - Do not quote or rewrite the full claim in the evaluation bullets.

\noindent - Evidence must be grounded in the video or scene graph; do not invent unsupported evidence.

\noindent - A claim is Critical if it concerns main subjects, primary actions/events, key attributes, key spatial relations, temporal order, or causal/event structure; otherwise, it is Minor.

\noindent - Judgment must be one of Supported, Partially Supported, or Contradicted. Use Contradicted when a required element is missing or conflicts with the evidence; use Partially Supported when the main requirement is present but incomplete or ambiguous.

\vspace{0.4em}
\noindent\textbf{Scoring scale:}

\noindent - 5 (Excellent): ALL claims (Critical and Minor) are Supported.

\noindent - 4 (Good): All Critical claims are Supported, or only 1--2 Critical claims are Partially Supported. At most 1--2 Minor claims are Contradicted.

\noindent - 3 (Fair): 1--2 Critical claims are Contradicted (e.g., missing specific event/object), but the core meaning remains recognizable.

\noindent - 2 (Poor): 3 or more Critical claims are Contradicted (e.g., core temporal sequence is broken), or only 1--2 Critical claims are Supported.

\noindent - 1 (Fail): ALL Critical claims are Contradicted. Any Supported or Partially Supported claims are strictly Minor.

\vspace{0.4em}
\noindent\textbf{[STEP 3 - Quality Assessment]}

\noindent Evaluate the intrinsic video quality along three dimensions through direct visual observation of the video. Wrap this step in \verb|<quality>...</quality>|.

\vspace{0.4em}
\noindent\textbf{Tasks:}

\noindent For each dimension, emit the given tag with a 2--3 sentence analysis and end with "Score: \verb|<1-5>|".

\noindent \verb|<vq>| Visual Quality - per-frame optical quality judged as still images, independent of motion. Check: noise/grain, blur, exposure, compression artifacts, color fidelity, resolution. \verb|</vq>|

\noindent \verb|<tq>| Temporal Quality - frame-to-frame stability. About stability, NOT amount of motion. Check: identity preservation, flickering, motion smoothness, background stability, clothing/accessory persistence. \verb|</tq>|

\noindent \verb|<pc>| Physical / Common-sense Consistency - adherence to real-world physics and common sense. Check: gravity, solid-body physics, anatomy, spatial relations, shadows/reflections, fluid physics, causality, contextual common sense. \verb|</pc>|

\vspace{0.4em}
\noindent\textbf{Rules:}

\noindent - Judge each dimension independently; do not let one dimension bias another.

\noindent - Base every claim on direct visual observation; do not reuse the scene graph.

\vspace{0.4em}
\noindent\textbf{Scoring scale:}

\noindent - 5 (Excellent): no noticeable issues detectable.

\noindent - 4 (Good): minor issues visible only on close inspection.

\noindent - 3 (Acceptable): issues noticeable during normal playback.

\noindent - 2 (Poor): issues dominate the viewing experience.

\noindent - 1 (Unacceptable): severely degraded or absent on this dimension.

\vspace{0.4em}
\noindent\textbf{[STEP 4 - Final Answer]}

\noindent Emit the final scores after closing \verb|</think>|.

\vspace{0.4em}
\noindent\textbf{Tasks:}

\noindent Emit exactly one line in the form:

\noindent \verb|<answer>SA <n> VQ <n> TQ <n> PC <n></answer>|

\vspace{0.4em}
\noindent\textbf{Rules:}

\noindent - Scores must match those written inside \verb|<think>|.

\end{tcolorbox}

\section{Qualitative Examples}
\label{app:qualitativeexamples}
\begin{table*}[t!]
\centering
\begin{minipage}{0.96\textwidth}
\footnotesize
\setlength{\parindent}{0pt}

\hrule
\vspace{0.5em}

This survey is conducted for the human evaluation of
\textit{Plan-and-Verify Video Reward Reasoning with Spatio-Temporal Scene Graph Grounding}.
The expected completion time is approximately 40 minutes.
Please carefully read the instructions before beginning the evaluation.

\vspace{0.4em}
\textbf{1) Evaluation Overview}

Given a text prompt and its corresponding video, the model produces a reasoning trace evaluating their alignment.
Your task is to judge whether this reasoning process is valid from a human perspective.

The evaluation consists of 16 examples, each containing a text prompt, a generated or natural video, and the model's reasoning trace.
The reasoning trace contains four components:
\textbf{Scene Graph (SG)}, which summarizes objects, attributes, and temporally grounded relationships in the video;
\textbf{Plan}, which decomposes the prompt into verification items;
\textbf{Analysis}, which provides evidence for each item and assigns \textit{Supported}, \textit{Partially Supported}, or \textit{Contradicted};
and \textbf{Final Analysis \& Score}, which provides an overall summary and a score from 1 to 5.

\vspace{0.4em}
\textbf{Scene Graph (SG) Format}

The scene graph is represented as a JSON object with
\texttt{objects} and \texttt{relationships}:

\begin{verbatim}
{
 "objects": [
  {"id":"0","category":"bird","attributes":["yellow"]},
  {"id":"1","category":"wooden post","attributes":["old"]}
 ],
 "relationships": [
  ["0","perching on","1",[[0,4]],"spatial"],
  ["0","flying","-1",[[4,6]],"action"]
 ]
}
\end{verbatim}

Each object is represented by an \texttt{id}, \texttt{category}, and \texttt{attributes}.
Each relationship follows \texttt{[subject id, predicate, object id, time interval, relation type]}, where the time interval indicates when the relation holds in seconds.
Object IDs start from ``0''; ``-1'' denotes the camera, and in relationships is used for camera motion or when there is no target object.
Relation types include \textit{spatial}, \textit{motion}, \textit{functional}, \textit{stateful}, \textit{social}, \textit{attentional}, and \textit{event-level} relations.

For example, the scene graph above means that ``a yellow bird (0) perches on an old wooden post (1) from 0--4 seconds and then flies away from 4--6 seconds.''

\vspace{0.4em}
\textbf{2) Evaluation Procedure}

For each example, read the reasoning trace and evaluate the following six criteria.
The response format differs across items, so follow the instructions provided for each item.
Importantly, you are \textbf{not} asked to directly judge whether the video matches the prompt;
instead, judge whether the model's reasoning process for evaluating the alignment is valid.

\vspace{0.25em}
\textbf{1. Scene Graph Accuracy.}
Do the objects, attributes, and relationships described in the scene graph actually appear in the video?
For example, if the scene graph states ``yellow bird,'' is there actually a yellow bird in the video?

\textbf{2. Plan Decomposition Quality.}
Are the prompt requirements properly decomposed into verification items without omissions or duplication?
Are any items not specified in the prompt introduced?

\textbf{3. Evidence Grounding.}
Can the evidence provided for each judgment actually be verified in the video or scene graph, rather than being fabricated?

\textbf{4. Claim-Scope Alignment.}
Is the provided evidence relevant to the specific verification item being evaluated?

\textbf{5. Judgment Validity.}
Does the judgment (\textit{Supported}, \textit{Partially Supported}, or \textit{Contradicted}) naturally follow from the provided evidence?

\textbf{6. Judgment Consistency.}
Are the judgments across different verification items logically consistent?

\vspace{0.5em}
\hrule
\end{minipage}

\caption{English translation of the instructions for the scene-graph and
reasoning-trace evaluation.
The original instructions were presented in Korean through the annotation interface.}
\label{tab:human_eval_instructions}
\end{table*}

\begin{figure*}[t!]
    \centering
    \includegraphics[width=\linewidth]{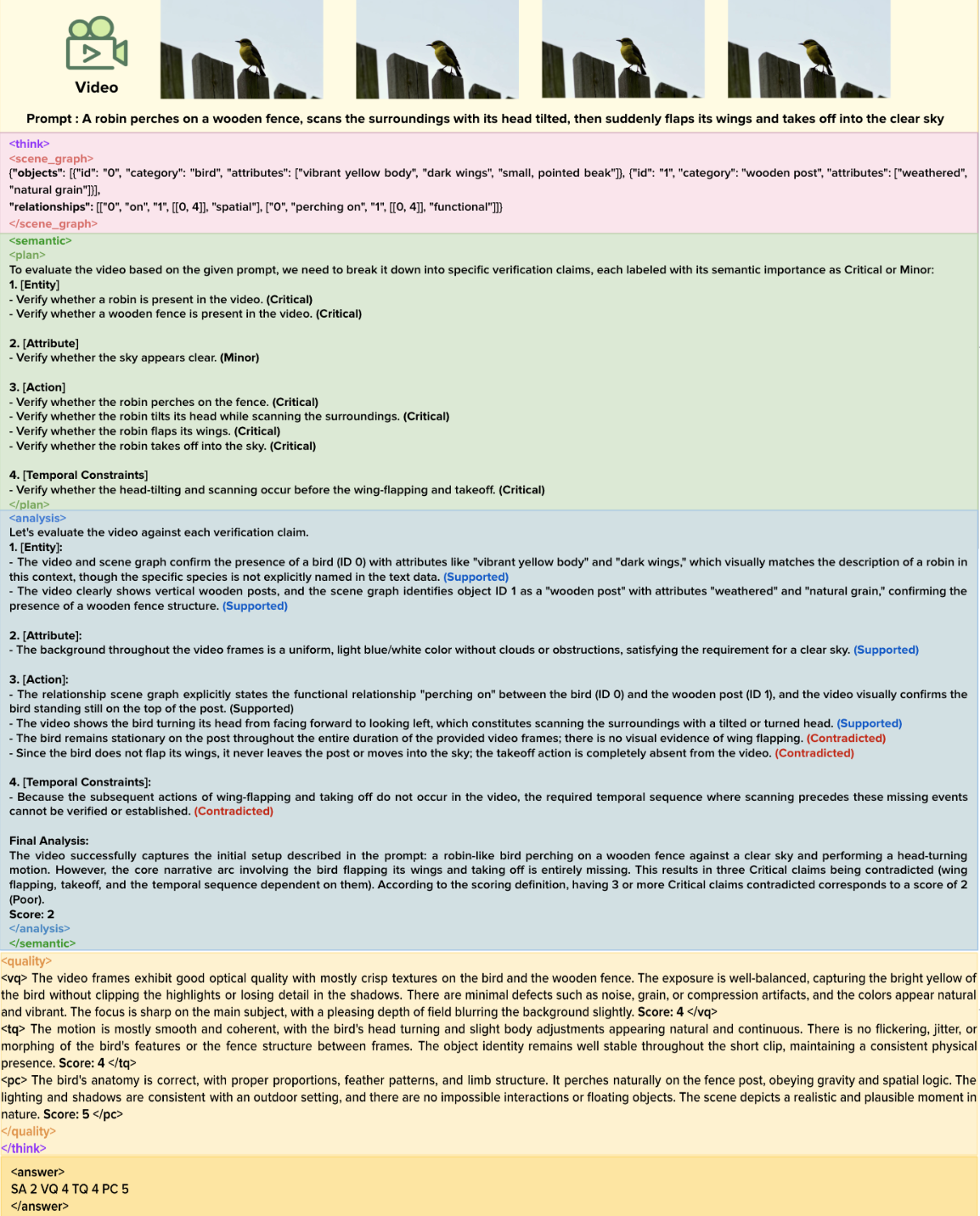}
    \caption{An example of SG-PVR's reasoning output on VideoScore-Bench-v2}
    \label{fig:model_output}
\end{figure*}

\begin{figure*}[t!]
    \centering
    \includegraphics[width=\linewidth, height=0.98\textheight, keepaspectratio]{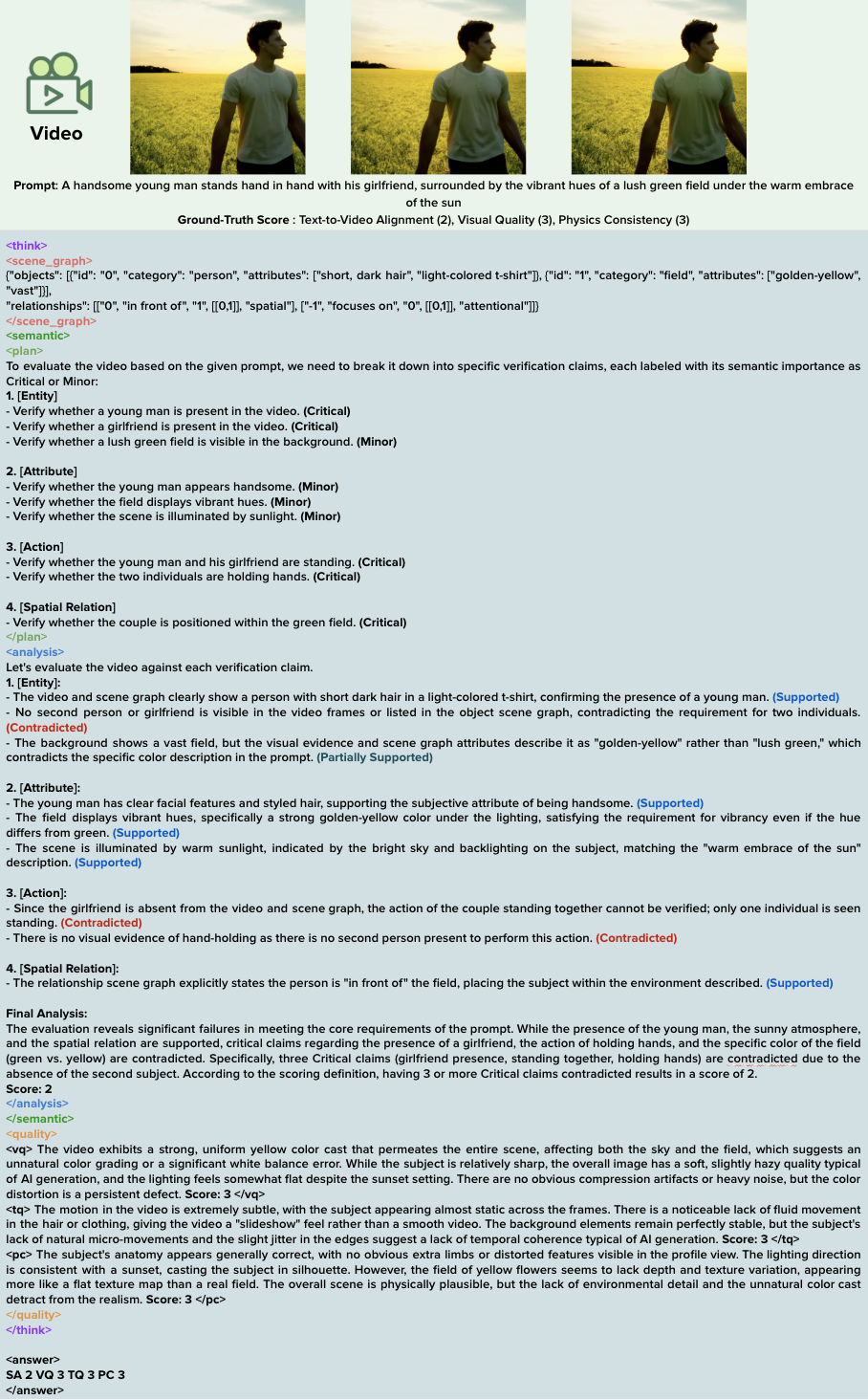}
    \caption{An example of SG-PVR's reasoning output on VideoScore-Bench-v2}
    \label{fig:model_output_partial}
\end{figure*}

Figures~\ref{fig:model_output} and \ref{fig:model_output_partial} show example reasoning outputs of SG-PVR on the VideoScore-Bench-v2. Specifically, in Figure \ref{fig:model_output}, given the prompt “A robin perches on a wooden fence, scans the surroundings with its head tilted, then suddenly flaps its wings and takes off into the clear sky” and the corresponding video, SG-PVR evaluates four aspects: semantic alignment, visual quality, temporal quality, and physical/common-sense consistency. 
The model first outputs a video scene graph in the \texttt{<scene\_graph>} tag. It then decomposes the prompt into atomic verification claims covering entities, attributes, actions, and temporal constraints in the \texttt{<plan>} tag. In the \texttt{<analysis>} tag, SG-PVR verifies each claim individually and provides a final analysis explaining the semantic alignment score. The \texttt{<quality>} tag contains separate evaluations for visual quality, temporal quality, and physics consistency. Finally, the \texttt{<answer>} tag summarizes all predicted scores.

% \begin{figure*}[t!]
%     \centering
%     \includegraphics[width=\linewidth]{figures/pairwise_compare.png}
%     \caption{Qualitative comparison of SG-PVR and Pairwise reward model(UnifiedReward-Think). SG-PVR's structured trace—scene graph, atomic claims, and per-claim verifications—provides explicit evidence for each judgment, while UnifiedReward-Think outputs holistic scores without visible grounding.}
%     \label{fig:pairwise_compare}
% \end{figure*}

% Figure~\ref{fig:pairwise_compare} compares SG-PVR with UnifiedReward-Think on a sample with the prompt "a cute raccoon playing guitar in a boat on the ocean". 
% UnifiedReward-Think outputs holistic per-dimension scores without grounding each in specific visual evidence and incorrectly prefers Video 1. 
% SG-PVR instead decomposes the prompt into atomic claims and verifies each separately—e.g., the spatial claim "inside the boat" is Partially Supported for Video 1 (no full boat hull visible) but Supported for Video 2 (raccoon clearly inside the hull)—and rates Video 2 higher across all four scores. 
% The structured trace makes each judgment auditable.

%기존
% \begin{figure*}[t!]
%     \centering
%     \includegraphics[width=\linewidth]{figures/pointwise_compare.png}
%     \caption{Qualitative comparison of SG-PVR and VideoScore2. SG-PVR's fine-grained claim verification identifies all missing prompt requirements, while VideoScore2's coarse evaluation overlooks several prompt-specific requirements}
%     \label{fig:pointwise_compare}
% \end{figure*}

\begin{figure*}[t!]
    \centering
    \includegraphics[width=\linewidth]{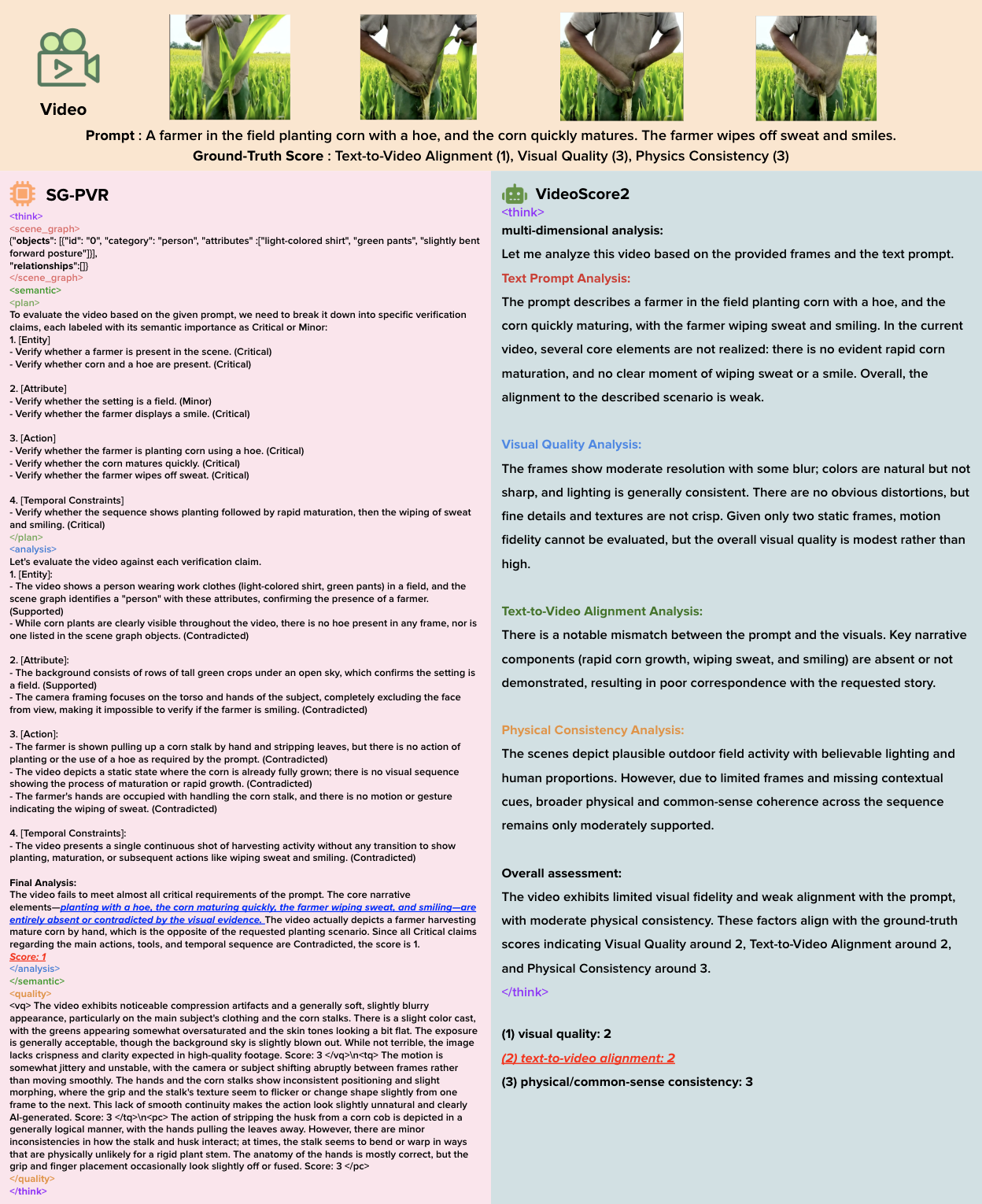}
    \caption{Qualitative comparison of SG-PVR and VideoScore2. SG-PVR's fine-grained claim verification identifies all missing prompt requirements, while VideoScore2's coarse evaluation overlooks several prompt-specific requirements.}
    \label{fig:pointwise_compare}
\end{figure*}

Figure~\ref{fig:pointwise_compare} compares SG-PVR with VideoScore2 on a sample with the prompt "A farmer in the field  planting corn with a hoe, and the corn quickly matures. The farmer wipes off sweat and smiles." 
Our model decomposes the prompt into atomic semantic claims and explicitly verifies each required entity, attribute, action, and temporal relation. VideoScore2, by contrast, fails to detect missing information (the absence of a hoe) and consequently produces an overestimated score.
% while VideoScore2 gives a coarse summary but does not verify each required action and temporal dependency separately. Also, Videoscore2 does not accurately assess the clarity of specific regions; it only evaluates overall clarity. Furthermore, it does not accurately detect violations of physics or common sense.

\begin{figure*}[t!]
    \centering
    \includegraphics[width=\linewidth]{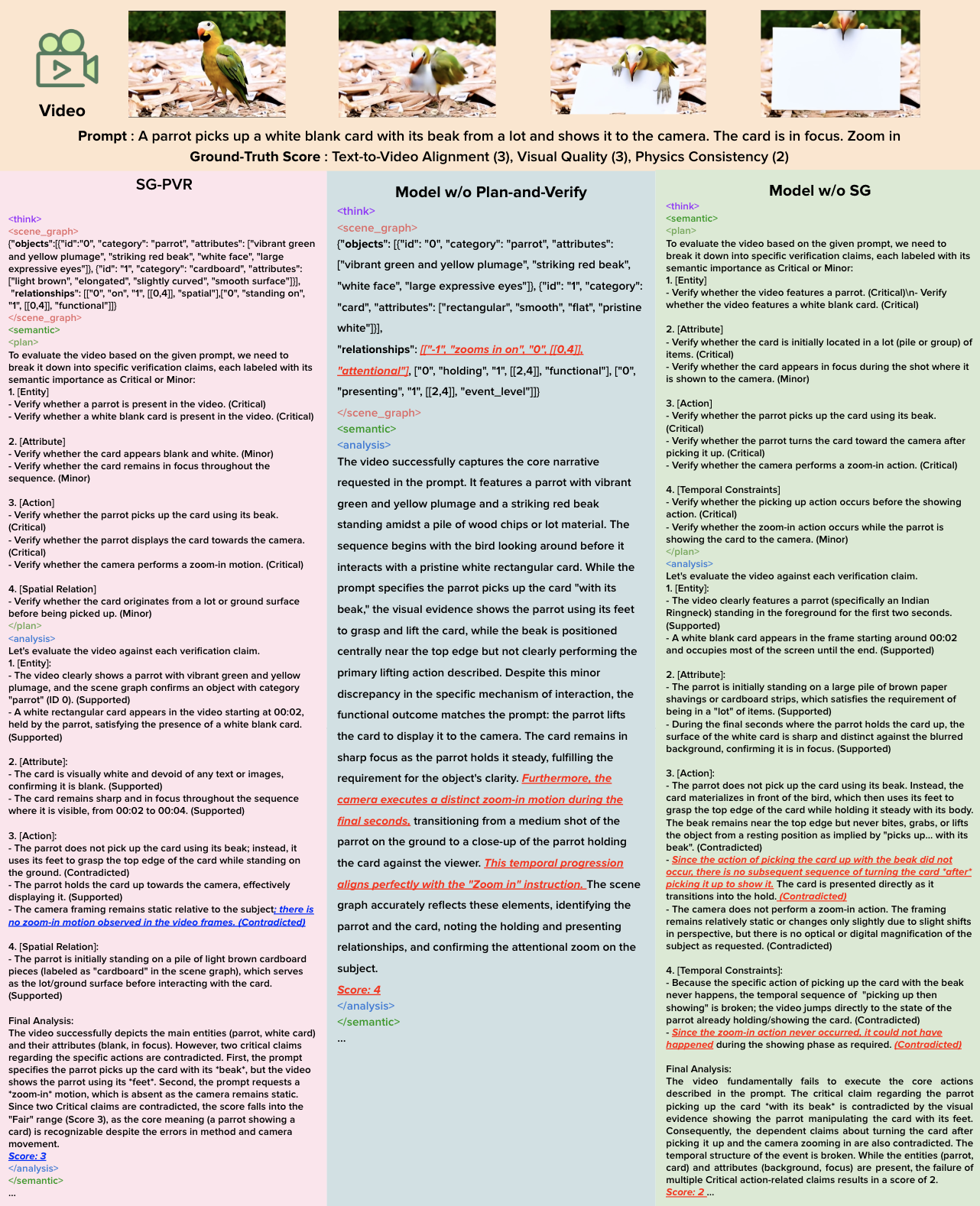}
    \caption{Qualitative comparison of SG-PVR against the w/o Plan-and-Verify and w/o SG variants. Unlike SG-PVR, which verifies each atomic claim with explicit scene graph evidence, the ablated variants produce inaccurate verification or under- or overestimated aggregate scores.}
    \label{fig:cmp_main_abl1_abl2_2}
\end{figure*}

Figure~\ref{fig:cmp_main_abl1_abl2_2} shows an example reasoning output of SG-PVR and its ablated variants on a prompt "A parrot picks up a white blank card with its beak from a lot and shows it to the camera. The card is in focus. Zoom in." 
Our model correctly identifies that no zoom-in occurred in the video by grounding the claim in explicit scene graph evidence. In contrast, the w/o SG variant verifies the same element multiple times across its reasoning, distorting the aggregated score through duplicate judgments, while the w/o Plan-and-Verify variant incorrectly judges that a zoom-in was performed—an error that is difficult to avoid without structured reasoning.

\end{document}